\documentclass{article}

\usepackage[final,sglblindworkshop]{neurips_2026}
\workshoptitle{World Models for High-Stakes Health}

\usepackage[utf8]{inputenc} 
\usepackage[T1]{fontenc}    
\usepackage{hyperref}       
\usepackage{url}            
\usepackage{booktabs}       
\usepackage{amsfonts}       
\usepackage{amsmath}
\usepackage{needspace}
\usepackage{float}
\usepackage{nicefrac}       
\usepackage{microtype}      
\usepackage{xcolor}         
\usepackage{graphicx}
\usepackage{subcaption}
\usepackage{float}
\newcommand{\repourl}{\url{https://github.com/edric-hao/Behavioral-Mirage-in-Medical-VLMs}}

\title{Beyond the Evidence Boundary: Behavioral Mirage Under Progressive Degradation in Medical VLMs}

\author{%
  Edric Castel C. Hao\thanks{Equal contribution.}\hspace{5pt}\thanks{This work does not relate to the author's position at Analog Devices, Inc.} \\
  Analog Devices, Inc. \\
  \texttt{edriccastel.hao@analog.com} \\
  \And
  Rez Samantha Z. Floresca$^{*}$ \\
  Ateneo de Manila Senior High School \\
  \texttt{rez.samantha.floresca@student.ateneo.edu} \\
}

\begin{document}

\maketitle

\begin{abstract}
VLMs for medical image interpretation are typically evaluated on clean inputs, despite the progressive evidence loss that can occur in high-stakes clinical settings. Existing robustness studies show that performance can degrade while confidence remains high, but do not establish when a previously reliable model-case pair ceases to behave reliably. We introduce a baseline-anchored framework that traces baseline-correct cases through progressive image degradation to identify an empirical evidence boundary, and define \emph{mirage} as continued commitment at this boundary, regardless of correctness. Across three VLMs on three medical datasets, only MedGemma provides a sufficiently large baseline-correct pool for trajectory analysis. On VQA-RAD, the only dataset in which the question text varies, language-only answerability is associated with baseline correctness (AUROC 0.677) and with commitment at the boundary (AUROC 0.653 to 0.670). MedGemma's post-boundary behavior is heterogeneous, including direct mirage, correct recovery, mirage dissolution, consistent abstention, and oscillation, with mirage rates of 16.0 to 100.0\% across dataset-modality pairs. Despite this behavioral heterogeneity, residual-stream probes decode commit-versus-abstain behavior at the boundary with AUROC of 0.895 to 0.994 under nested cross-validation. These results show that progressive evidence loss can produce persistent commitment that output entropy and abstention behavior do not consistently reflect, motivating boundary-aware evaluation, option-order controls, and improved abstention mechanisms for reliable medical foundation models operating under partial observation.
\end{abstract}

\textbf{Introduction}

Vision-language models (VLMs) are increasingly used for medical image interpretation and clinical decision support~\citep{li2023llavamed,sellergren2025medgemma}. Yet their reliability is evaluated predominantly on complete, clean inputs, despite the progressive evidence loss that can arise from limited imaging hardware, acquisition conditions, storage, transmission, and computational infrastructure in low-resource settings. A model may appear reliable under standard evaluation, yet pose severe safety risks when deployed in high-stakes clinical settings, where complete visual evidence is not always guaranteed.

Existing robustness evaluations in medical VLMs measure performance across corruption severities. MedQ-Deg~\citep{liu2026medqdeg}, for example, demonstrates that model accuracy drops sharply under degradation even while confidence remains high. Such evaluations establish whether performance deteriorates, but not when a previously reliable model-case pair stops behaving reliably. Nor do they characterize how model behavior changes once the available evidence becomes insufficient to support its prediction. This distinction is important for validating medical foundation models, where reliability depends not only on predictive performance but also on whether model confidence and decisions remain appropriate to the evidence available to the system.

This gap matters because an optimal medical VLM should recognize when available evidence is insufficient and abstain rather than produce an unsupported clinical commitment. While studies on image removal demonstrate that VLMs remain overconfident even when medical images are completely omitted~\citep{asadi2026mirage,benlevi2026mirageprobes,medvigil2026broken}, existing work rarely evaluates abstention as a comprehensive safety mechanism. Prior studies also fail to establish a baseline of reliable behavior before evidence loss occurs, nor do they track how abstention should dynamically respond across progressive levels of evidence degradation. Without this, correct answers on degraded inputs cannot be distinguished from evidence-grounded behavior, ungrounded guessing, parametric priors, or shortcut learning~\citep{tong2024eyeswideshut,rubi2019,si2022languagepriornotonly,radimagenetvqa2026,medbiasx2025,decoct2025}.

This study therefore proposes a framework for systematically addressing these gaps by (1) establishing a baseline of high-confidence correct behavior, (2) determining when and how this reliability breaks down under progressive image degradation by operationalizing an empirical evidence boundary from observable behavioral changes, and examining how prompt-language degradation affects behavior, (3) characterizing what follows from such failures and the resulting behavioral trajectories, (4) analyzing how the presence of an abstention option influences model behavior under degradation, (5) redefining mirage behavior relative to the boundary, and (6) probing whether the model's internal representations separate abstention from commitment. Before such models can be trusted as components of larger clinical reasoning or patient world-model systems, it is necessary to establish when their predictions remain grounded in available evidence, when they should abstain, and how they behave once that evidence becomes insufficient. Together, these analyses characterize how medical VLM behavior changes as visual or linguistic evidence is degraded. 

\textbf{Related Work}

Robustness evaluation in medical vision-language models initially focused on specific modalities and targeted tasks, such as corruption resilience in digital pathology or out-of-distribution performance in MRI segmentation. MedQ-Deg~\citep{liu2026medqdeg} expanded this paradigm by evaluating 40 Medical VLMs across radiologist-calibrated degradation types spanning seven imaging modalities. Its central finding is what the authors term an \emph{AI Dunning-Kruger effect}: as accuracy collapses under increasing degradation, model confidence remains high or even increases across both severity levels and models of differing capability. While this finding directly motivates our work, MedQ-Deg measures confidence via output-label voting entropy at isolated severity levels. It neither tracks individual cases through degradation trajectories to identify when reliable behavior breaks down and what follows, nor evaluates behavior when models can abstain rather than make a forced choice. These limitations are important for medical foundation models intended to support reliable reasoning in high-stakes settings, where observations may be incomplete or progressively degraded.

Beyond single-response accuracy, prior work has used repeated sampling to characterize model consistency and uncertainty. Self-consistency approaches sample multiple responses to the same question and use the resulting answer distribution to quantify uncertainty, with greater agreement corresponding to lower response entropy~\citep{wang2022selfconsistency,taubenfeld2025confidence}. Repeated sampling therefore provides a stronger basis for establishing reliable behavior by identifying cases in which models respond consistently and correctly, rather than treating a single correct response as sufficient evidence. However, existing work has not examined how model behavior changes before, during, and after reliability breaks down under progressive image degradation in medical VLMs. This leaves open whether response consistency remains informative as the evidence available to the model progressively deteriorates.

\citet{benlevi2026mirageprobes} define mirage as output alignment between image-present and image-absent conditions without first establishing whether the baseline image-present response is correct. Models relying on strong priors or dataset biases can therefore produce identical, ``aligned'' answers with or without the image by chance. This fails to distinguish genuine visual evidence integration from a model whose prior happens to align with the ground truth. We instead define mirage relative to the population-level breaking threshold, defined in the methodology. Once the threshold is reached, continued commitment is classified as mirage regardless of whether the output is correct or incorrect. This reframes mirage as a boundary-level reliability behavior, providing a direct way to evaluate whether model commitments remain supported as available evidence becomes insufficient.

While selective prediction and abstention have long been studied in machine learning, encouraging LLMs and VLMs to withhold answers under uncertainty remains challenging because abstention does not necessarily indicate a lack of knowledge. \citet{srinivasan2024selectiveselectivepredictionreducing} show that VLMs may over-abstain on questions they could answer correctly, while models may also answer rather than abstain when uncertain, reflecting overconfident guessing~\citep{liu2025abstain,aravindan2026secondguess}. Examining abstention alongside forced responses may therefore reveal whether recurring answers reflect uncertainty or systematic response patterns and biases that emerge when models cannot express uncertainty through abstention. This distinction is particularly relevant to high-stakes medical foundation models, where appropriate abstention may be necessary when available evidence is inadequate.

Finally, interpretability work asks whether such decisions may be detectable before an answer is generated. \citet{morenocencerrado2025noanswer} show that forthcoming answer correctness can be linearly decoded from question-only residual-stream activations across multiple LLMs, with spontaneous abstentions clustering at the negative extreme of the learned correctness direction. They extract question-only residual-stream activations and fit a difference-of-means linear direction that predicts factual correctness across multiple LLMs, generalizing out-of-distribution while failing to transfer to arithmetic reasoning~\citep{morenocencerrado2025noanswer}. This leaves open whether abstain-versus-commit is itself a directly probeable representational axis or merely a downstream correlate of correctness. Determining whether such a signal exists would connect behavioral evidence of appropriate uncertainty expression to an interpretable internal mechanism.

Taken together, these existing works and gaps motivate and justify our methodological approach.

\textbf{Methodology}

\vspace{2pt}
\noindent
\begin{minipage}{\linewidth}
\centering
\includegraphics[width=0.90\linewidth, trim=0 40pt 0 0, clip]{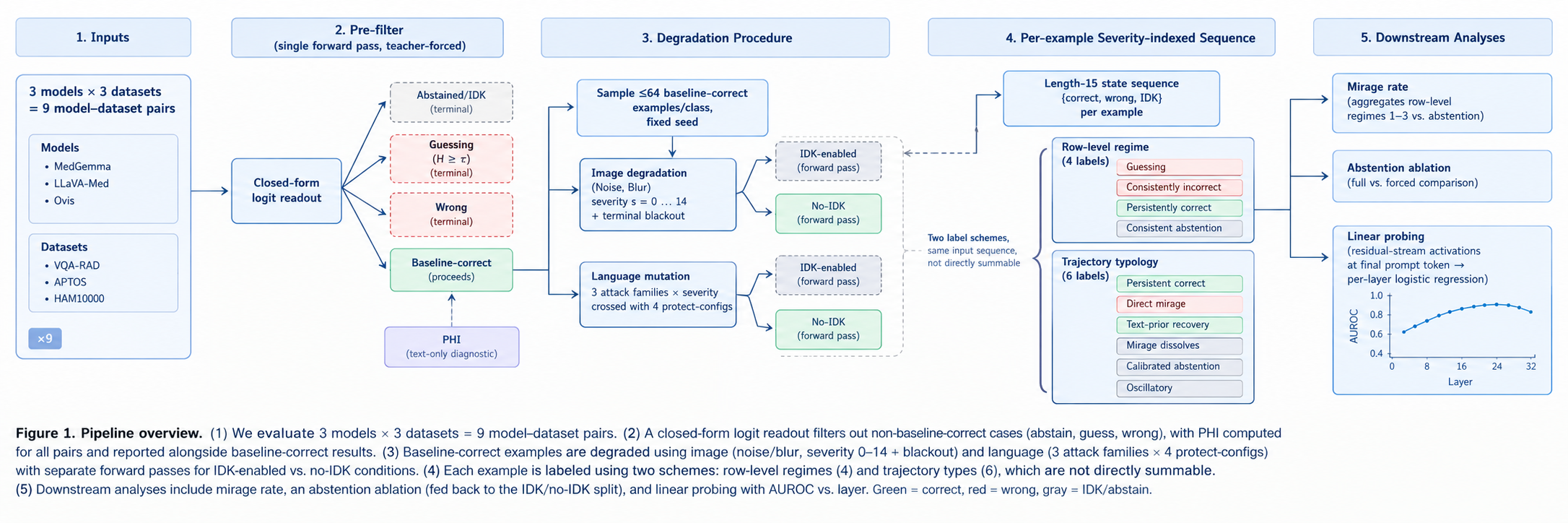}
\vspace{-2pt}
\captionof{figure}{Schematic overview of the proposed evaluation framework and diagnostic pipeline.}
\label{fig:overview}
\end{minipage}
\vspace{6pt}

\textbf{Models and Datasets}

We screen three VLMs spanning medical and general-domain models (\textbf{MedGemma}~\citep{sellergren2025medgemma}, \textbf{LLaVA-Med}~\citep{li2023llavamed}, and \textbf{Ovis}~\citep{lu2025ovis25}) against three medical visual question answering and classification datasets (\textbf{VQA-RAD}~\citep{lau2018vqarad}, restricted to its closed-ended yes/no subset, \textbf{APTOS}~\citep{aptos2019}, and \textbf{HAM10000}~\citep{tschandl2018ham10000}), yielding nine model-dataset pairs. We apply the same pipeline of pre-filtering, degradation, abstention ablation, PHI diagnostics, and linear probing across these model-dataset pairs, with downstream analyses run where enough baseline-correct examples exist. Each dataset is framed as a closed-ended task with a designated set of substantive choices alongside a single abstention choice. This gives a common answer space for tracking substantive commitment and abstention across models and datasets. Option letters are fixed across experiments: APTOS uses A (No DR), B (Mild DR), C (Moderate DR), D (Severe DR), E (Proliferative DR), and F (abstention). HAM10000 uses A (actinic keratosis), B (basal cell carcinoma), C (benign keratosis), D (dermatofibroma), E (melanoma), F (melanocytic nevus), G (vascular lesion), and H (abstention). VQA-RAD uses A (Yes), B (No), and C (abstention).

\textbf{Pre-filter Mechanism}

Each item is posed as a closed-ended completion over its dataset's fixed answer set, with explicit instruction to select the abstention choice when the model cannot answer with confidence (the IDK-enabled condition). We obtain the predicted answer directly from the output distribution via a teacher-forced, single-forward-pass readout: the image and prompt populate the key-value cache once, a fixed answer-prefix string is force-decoded, and we extract the logits at the position immediately following the prefix, restricted to the valid answer tokens. Let $z \in \mathbb{R}^K$ be these logits over the $K$ answer choices. The predicted answer is $\hat{y} = \arg\max_k z_k$, and uncertainty is the Shannon entropy of the softmax distribution,
{\small\[
\textstyle H = -\sum_{k=1}^{K} p_k \ln p_k, \qquad p = \mathrm{softmax}(z).
\]}%
Because the readout is deterministic, a single forward pass characterizes baseline behavior. Each example is assigned to one of four baseline categories: \emph{abstained/IDK} ($\hat{y}$ is the abstention choice), \emph{high-entropy commitment} (not abstained, $H \geq \tau$), \emph{baseline-correct} (not abstained, $H < \tau$, $\hat{y}$ correct), or \emph{wrong} (not abstained, $H < \tau$, $\hat{y}$ incorrect). We set $\tau = 0.30$ nats, consistent with entropy-based rejection thresholds used in related VLM uncertainty-quantification work under prompted abstention~\citep{wang2025vision}. Only \emph{baseline-correct} examples proceed to degradation, defining a clean-input reference population that is both correct and concentrated in its predicted answer. All downstream trajectory, mirage-rate, and regime analyses are therefore conditional on membership in the baseline-correct population and do not represent unconditional model failure rates.

\textbf{Language-Only Answerability (PHI)}

Medical vision-language tasks are vulnerable to language priors, where a model predicts the ground-truth answer from question text alone without relying on visual evidence. We therefore compute a text-only, image-free diagnostic score, the Prior Harnessing Index (\textbf{PHI}) of \citet{benlevi2026mirageprobes}, for every item across all datasets before degradation. No image is passed in either condition. For a given item, we score the log-probability assigned to the ground-truth answer choice conditioned on (a) the item's actual question text and (b) a fixed ``null'' question (``What is the answer?'' plus the dataset's options), using the same single-position logit readout. Here the log-probability is that of the ground-truth option letter under the softmax over option letters, rather than a free-form answer likelihood.
{\small\[ \textstyle \mathrm{PHI} = \log P(\mathrm{answer} \mid \mathrm{question}, \mathrm{no\ image}) - \log P(\mathrm{answer} \mid \mathrm{null\ question}, \mathrm{no\ image}). \]}%
A high, positive PHI indicates that the ground-truth answer is more predictable from question language alone relative to the null prompt, consistent with stronger language-based answerability. On APTOS and HAM10000 the prompt is a fixed per-dataset template, so PHI there reflects the task instruction and not item-specific question content, and we restrict interpretation of question effects to VQA-RAD. An image-present variant, which keeps the clean image in both prompts, is reported for comparison in Appendix~D.

\textbf{Degradation Procedure}
For every model-dataset pair, up to 64 baseline-correct examples per ground-truth class are sampled using a fixed seed without replacement. This post-filter cap reduces the dominance of classes with disproportionately large baseline-correct pools. Classes with fewer than 64 available examples are sampled in full, while classes with fewer than 16 available examples are excluded (APTOS grade B has one baseline-correct example and HAM10000 dermatofibroma has none). The resulting aggregate estimates are therefore based on the class-balanced sample rather than the datasets' natural class frequencies. Sampled examples are then passed through the full severity schedule described below. Both image and language degradation reuse the same forced-choice logit readout mechanism as the pre-filter. Readouts are recorded at every level of a fixed, predefined severity schedule rather than through adaptive early stopping, preserving each example's complete response trajectory across the evaluated severities. At each severity level, examples are evaluated under two conditions: an \emph{IDK-enabled} condition containing the explicit abstention choice and prompt instruction, and a \emph{no-IDK} condition omitting both.

\textbf{Image Degradation: Noise and Blur}

Visual evidence is degraded independently using two corruption families, with language held fixed at its clean form. Both use a shared severity index $s \in \{0, 1, \dots, 14\}$, where $s=0$ denotes clean input and $s=14$ denotes terminal blackout, in which the image is replaced by an all-black image (Appendix Table~\ref{tab:severity-schedule}). The shared index is ordinal and does not equate noise magnitude with blur width. Degradations are applied to the image at its native resolution, before the model's own resizing.

Noise is zero-mean Gaussian with standard deviation $c_s$ on the unit intensity scale, clipped to $[0,1]$, and blur is an isotropic Gaussian kernel of standard deviation $\rho_s$ pixels. The coefficients $c_s$ and $\rho_s$ for each $s$ are listed in Appendix~L.

\textbf{Language Mutation Types and Severity}

Prompt-language information is degraded with the image held fixed at its clean form, using WordNet synonym substitution and random character corruption at three rates each and blackout masking at a single full rate, under four protection configurations (\texttt{protect\_all}, \texttt{protect\_idk}, \texttt{protect\_desc}, and \texttt{protect\_none}) that shield the abstention instruction, the answer-choice descriptions, both, or neither, with the opening and closing lines always mutated. For VQA-RAD only the question text is mutated. Full definitions are in Appendix~K.

\textbf{Population-Level Breaking Threshold}

The population-level breaking threshold $s^\ast$ is the first evaluated severity at which the share of \emph{confident-correct} examples (answered, $H<\tau$, correct) in the evaluated sample falls to or below $p_{\max}$, the largest class share of that same class-balanced sample. A constant predictor that always outputs the largest class would score $p_{\max}$ on this sample, so $s^\ast$ marks the first severity at which the aggregate is consistent with fixed-answer collapse. It does not show that visual evidence has disappeared or that individual predictions are evidence-independent. Because correctness need not decrease monotonically, $s^\ast$ is the first crossing and not an irreversible transition. If the share never reaches $p_{\max}$, no threshold is assigned. The choice of statistic and reference matters, and Appendix~F reports $s^\ast$ and mirage rates under alternative choices, with bootstrap intervals.

\textbf{Post-Boundary Behavioral Trajectories Across Severity}

To capture trajectory shape, we encode each baseline-correct example as a length-15 sequence of states from $\{\emph{correct}, \emph{wrong}, \emph{IDK/abstain}\}$ across severities $s=0,\dots,14$, with $s=14$ corresponding to blackout. Low-confidence wrong predictions ($\max_k p_k < 0.5$, $\hat{y} \neq$ answer) are pooled into \emph{wrong}, since the model still commits to an incorrect answer. For correct predictions, low-confidence cases ($\max_k p_k < 0.5$, $\hat{y}=$ answer) are instead treated as IDK-like states because the correct answer is the argmax but does not receive a majority of the predicted probability mass. This asymmetry reflects that a wrong commitment is a commitment, whereas a low-confidence correct answer is not a confident success, and it does not imply that the model explicitly abstained. Appendix~G reports counts when low-confidence wrong answers are also treated as IDK. Thus, IDK/abstention for this typology is defined as $\mathrm{abstained}=\mathrm{True}$ \emph{or} ($\max_k p_k < 0.5$ and $\hat{y}=$ answer). This criterion is coarser than the $\tau=0.30$-nat entropy threshold used for regime classification, so labels can differ. We retain the entropy criterion for regime classification (Appendix Table~\ref{tab:regimes}) because it accounts for the full answer distribution, while the $0.5$ maximum-probability criterion is used only for trajectory abstraction.

Each example's full sequence is assigned to exactly one of six mutually exclusive types by a fixed precedence rule: persistent correct, direct mirage, correct recovery, mirage dissolves, consistent abstention, and oscillatory. The definitions are in Appendix~A. Because the blackout image is identical for every example, Appendix~G also reports the counts with the blackout step removed.

Ground-truth and error-choice proportions at each severity are defined in Appendix~B.

\textbf{Mirage Definition and Mirage Rate}

We define \emph{operational mirage behavior} as continued substantive answering at the population-level breaking threshold $s^\ast$, regardless of whether the answer is correct or incorrect. Thus, an example may remain correct while still exhibiting operational mirage behavior at the threshold. This label describes behavior at the population reliability boundary and does not by itself establish that the prediction is unsupported, hallucinated, or independent of visual evidence.

At the population-level breaking threshold $s^\ast$, we report the \textbf{post-boundary substantive commitment rate}, defined as the proportion of evaluated baseline-correct examples that produce a substantive commitment rather than abstaining. We report this rate per model, dataset, and degradation family as the operational measure of mirage prevalence.

\textbf{Linear Probing}

We probe whether abstain-versus-commit behavior is linearly decodable from the model's residual-stream activations at each model-dataset-degradation pair's population-level breaking threshold $s^\ast$. For each example, we extract the residual-stream activation $h_\ell \in \mathbb{R}^d$ at the final token of the chat-template generation prompt (which precedes the forced answer prefix) of layer $\ell$ from a single deterministic forward pass without generation. Examples are labeled by the model's own output at $s^\ast$: \textbf{commit} ($y=1$) if the model produces an answer and \textbf{abstain} ($y=0$) if it produces ``I don't know.'' Blur layer curves use balanced subsamples. For each layer $\ell$, we standardize $h_\ell$ within each training fold and fit an $\ell_2$-regularized, class-balanced logistic regression probe $f_\ell: \mathbb{R}^d \to [0,1]$ using 5-fold stratified cross-validation. We report mean AUROC as the layer's separability score and select the best-separating layer as $\ell^\star = \arg\max_\ell \mathrm{AUROC}_\ell$. The class-only baseline is the same procedure applied to a one-hot encoding of the ground-truth class. Because $\ell^\star$ is chosen on the same folds that score it, we also report nested cross-validated AUROC, with the layer and regularization strength chosen inside the training folds, at four token positions and from the clean input (Appendix~J).

\textbf{Results and Discussion}

\textbf{Prefilter Outcomes}

\vspace{1pt}
\noindent
\begin{minipage}{\linewidth}
\centering
\tiny
\renewcommand{\arraystretch}{0.75}
\setlength{\tabcolsep}{1.5pt}
\captionof{table}{Prefilter outcomes by model-dataset pair, in percent of the $N$ items. IDK is abstention, High-ent. is a high-entropy commitment, and BC is baseline-correct.}
\label{tab:prefilter}
\vspace{1pt}
\begin{tabular}{llccccc}
\toprule
\textbf{Model} & \textbf{Dataset} & $N$ & \textbf{IDK\%} & \textbf{High-ent.\%} & \textbf{BC\%} & \textbf{Wrong\%}\\
\midrule
MedGemma & APTOS & 2930 & 2.9 & 16.4 & 55.4 & 25.3 \\
MedGemma & HAM10000 & 10015 & 0.0 & 47.0 & 33.1 & 20.0 \\
MedGemma & VQA-RAD & 1191 & 0.1 & 17.0 & 58.9 & 24.0 \\
LLaVA-Med & APTOS & 2930 & 0.0 & 100.0 & 0.0 & 0.0 \\
LLaVA-Med & HAM10000 & 10015 & 0.0 & 100.0 & 0.0 & 0.0 \\
LLaVA-Med & VQA-RAD & 1191 & 0.0 & 97.6 & 2.0 & 0.4 \\
Ovis & APTOS & 2930 & 100.0 & 0.0 & 0.0 & 0.0 \\
Ovis & HAM10000 & 10015 & 0.0 & 100.0 & 0.0 & 0.0 \\
Ovis & VQA-RAD & 1191 & 98.6 & 1.4 & 0.0 & 0.0 \\
\bottomrule
\end{tabular}
\end{minipage}
\vspace{1pt}

Pool sizes are uneven across model-dataset pairs. MedGemma has a usable baseline-correct (BC) pool on all three datasets, covering 33.1 to 58.9\% of examples. By contrast, LLaVA-Med meets the minimum pool criterion only on VQA-RAD ($n=24$), while Ovis meets it on none. We therefore restrict downstream trajectory and regime analyses to MedGemma, and all tables below refer to MedGemma. For MedGemma, we sample up to 64 BC examples per ground-truth class, yielding $n=256$ for APTOS, $n=346$ for HAM10000, and $n=128$ for VQA-RAD. The HAM10000 sample is not fully balanced because two classes contain only 61 and 29 BC examples. LLaVA-Med/VQA-RAD ($n=24$) is retained for descriptive analysis only because its small pool does not support stable trajectory- or regime-level statistics.

On APTOS, LLaVA-Med assigns all items to high-entropy commitment, while Ovis abstains on all items. LLaVA-Med shows the same high-entropy pattern on HAM10000, and Ovis abstains on 98.6\% of VQA-RAD items but on none of the HAM10000 items under the same prompt template. To test whether these outcomes come from the readout, we ran 119 to 120 items per pair under the original and reversed option order (rotated for MedGemma) and compared the readout letter with a parsed free-generation answer (Appendix Table~\ref{tab:readout-val}). For MedGemma the readout agrees with free generation on 96.6 to 100\% of items. Ovis abstains under both orders on APTOS and VQA-RAD, but returns letter A under both orders on HAM10000, so its HAM10000 output is a fixed letter. LLaVA-Med returns the same displayed letter for every item in five of its six dataset-order pairs. For these two models the free-generation check cannot confirm the readout, because both outputs are nearly constant and agree trivially. However, it exposes a mismatch for LLaVA-Med on APTOS in the original order, where the readout answers A for every item but free generation says mild DR for 112 of 120. These outcomes therefore appear to be model behavior under this prompt and not only a readout artifact, although a prompt-format mismatch is not excluded. A larger $\tau$ would admit LLaVA-Med items, but because it returns the same displayed letter for nearly every image, the admitted items would be correct mainly by coincidence between that letter and the label. This differs from \citet{benlevi2026mirageprobes}, where Ovis produced enough substantive responses on VQA-RAD to build a 235-example mirage dataset without an explicit abstention option. The setups differ in prompt format and VQA-RAD subset, but the contrast is consistent with commitment rates depending on whether an IDK option exists.

\textbf{The Evidence Boundary by Dataset and Modality}

\vspace{-2pt}

\begin{figure}[H]
\centering

\subsubsection*{M.3 Class-Level IDK-Rate Heatmaps}

\begin{subfigure}[b]{0.48\textwidth}
  \includegraphics[width=\linewidth]{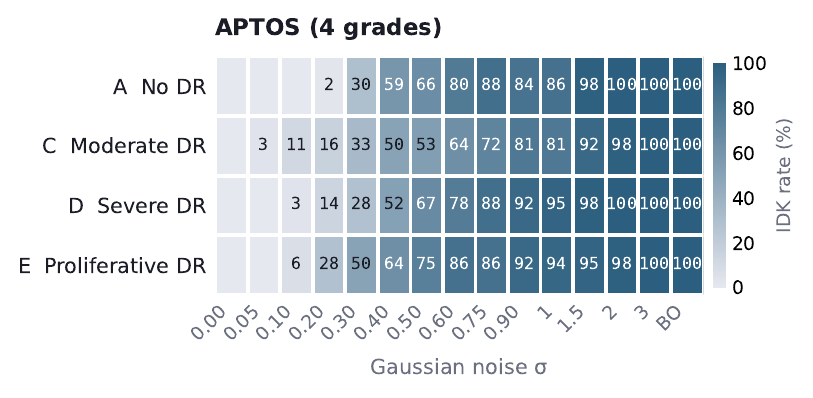}
  \caption{APTOS, noise}
\end{subfigure}
\hfill
\begin{subfigure}[b]{0.48\textwidth}
  \includegraphics[width=\linewidth]{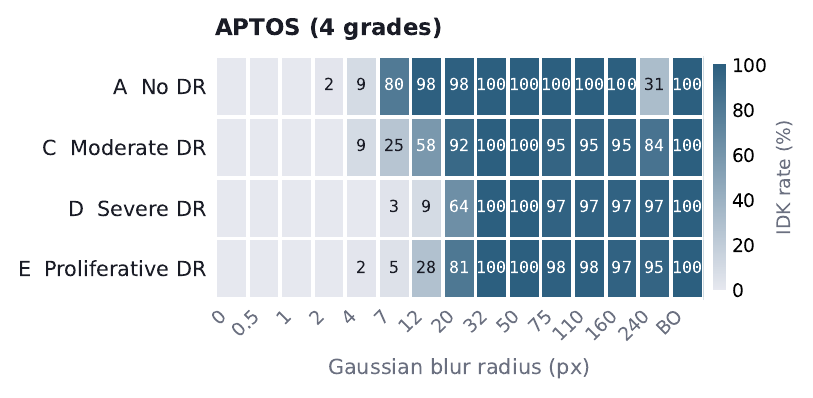}
  \caption{APTOS, blur}
\end{subfigure}

\vspace{0.5em}

\begin{subfigure}[b]{0.48\textwidth}
  \includegraphics[width=\linewidth]{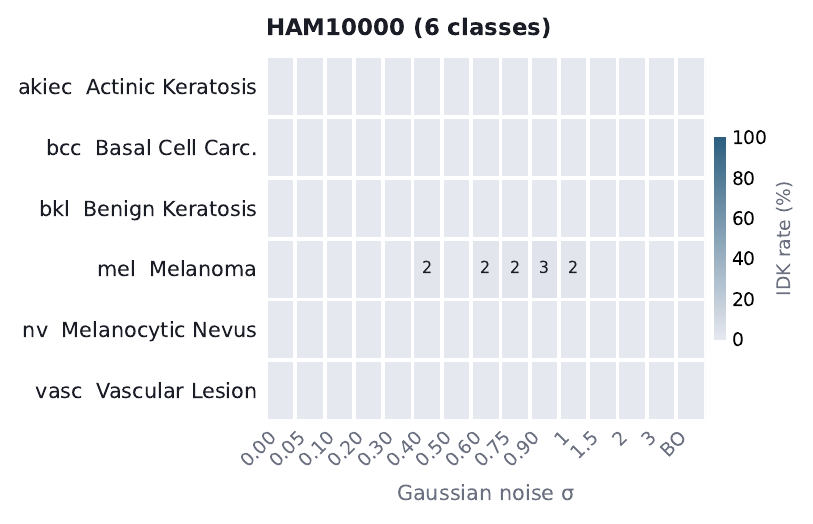}
  \caption{HAM10000, noise}
\end{subfigure}
\hfill
\begin{subfigure}[b]{0.48\textwidth}
  \includegraphics[width=\linewidth]{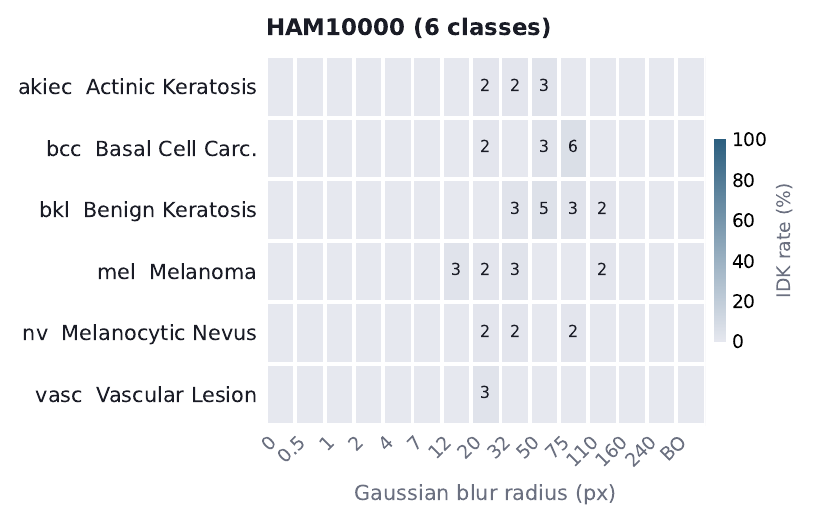}
  \caption{HAM10000, blur}
\end{subfigure}

\vspace{0.5em}

\begin{subfigure}[b]{0.48\textwidth}
  \includegraphics[width=\linewidth]{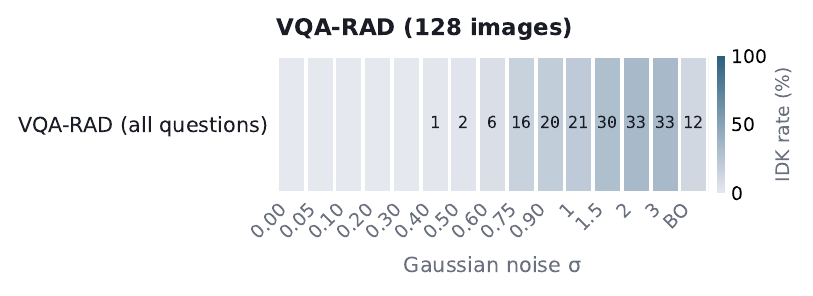}
  \caption{VQA-RAD, noise}
\end{subfigure}
\hfill
\begin{subfigure}[b]{0.48\textwidth}
  \includegraphics[width=\linewidth]{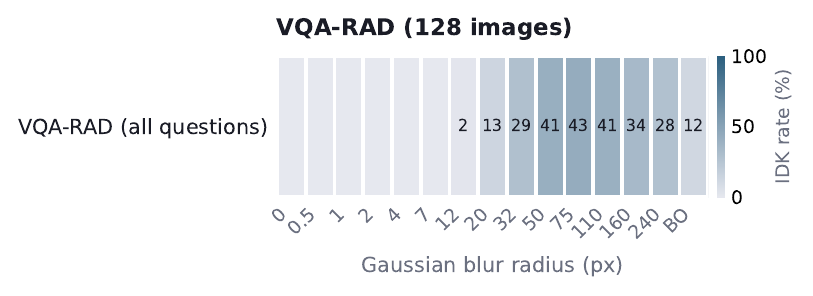}
  \caption{VQA-RAD, blur}
\end{subfigure}

\caption{IDK (abstention) rate by ground-truth class and severity for MedGemma across each dataset-modality pair. Color encodes the percentage of examples in that class abstaining at that severity level, and labels are rounded to integers. HAM10000 panels show a uniformly low IDK rate across all classes and severities, consistent with the main text's finding that HAM10000 almost never abstains under image degradation (see ``Class-level trajectories'' in Appendix~K). APTOS and VQA-RAD panels show class-dependent onset of abstention, complementing the aggregate class-level breakdowns discussed in the main text.}
\label{fig:idk_heatmaps}
\end{figure}

Breaking severities differ across datasets and modalities: the median is earliest on HAM10000 (2 for noise, 5 for blur), followed by APTOS (4, 6) and VQA-RAD (9, 8) (Appendix Table~\ref{tab:boundary-summary}). Noise produces an earlier boundary than blur on APTOS and HAM10000, while this ordering reverses on VQA-RAD. The boundary is also not universal: no APTOS or HAM10000 example remains correct through full blackout, whereas 37/128 (28.9\%) of VQA-RAD examples under noise and 32/128 (25.0\%) under blur remain correct at blackout. These are the \emph{Persistent correct} counts in Appendix Table~\ref{tab:trajectory-breakdown}. However, persistent correctness through a black image is not a robustness result, because it shows that the answer is recoverable without the image, and 34 of the 37 and 28 of the 32 such examples  have the correct answer ``No''. The relationship between degradation and abstention also varies by dataset and modality (Figure~\ref{fig:idk_heatmaps} and Appendix Table~\ref{tab:propensity}). Notably, HAM10000 almost never abstains: 35 of 9,688 degraded outputs (0.4\%) select the abstention option. Removing the abstention option at blackout (Appendices~C and~L) leaves HAM10000 accuracy at 18.5\% but drops APTOS to 25.0\% with a fixed letter.

\textbf{Behavioral Trajectories}

The trajectory counts (states defined in Appendix~A) depend on the blackout step, whose input carries no image information. At blackout MedGemma answers A for all 346 HAM10000 examples, abstains on all 256 APTOS examples, and on VQA-RAD answers No for 81, Yes for 32, and abstains for 15 of 128. A recovery at this step therefore reflects agreement between the prompt-driven default and the ground truth, not recovery of image-based reasoning. On VQA-RAD, correct recovery is the largest trajectory under both modalities (42/128 noise, 47/128 blur), followed by persistent correctness (37, 32) and direct mirage (27, 20). Without the blackout step, recoveries fall to 2 (noise) and 20 (blur). On HAM10000, all 64 recoveries are the 64 actinic keratosis (class A) examples, whose ground truth coincides with the constant output. Without the blackout step, the dataset is mostly direct mirage (251 and 258 of 346). APTOS is dominated by trajectories that terminate in abstention, particularly mirage-dissolve and consistent abstention, whereas HAM10000 is dominated by non-abstaining trajectories, including direct mirage, correct recovery, and oscillatory behavior. This pattern is consistent with the broader dataset-level variation in abstention behavior. Appendix~G gives the full counts.

\textbf{Post-Boundary Commitments: Ground Truth, Error, and Abstention}

At each pair's population-level breaking threshold $s^\ast$, defined as the first severity at which aggregate correctness falls to or below $A_{\mathrm{null}}=p_{\max}$, we decompose the baseline-correct population into ground-truth correct, non-ground-truth commitments, and abstentions. We further break down non-ground-truth commitments by predicted answer class (Appendix Table~\ref{tab:gt-error-sstar}, with selected-severity trajectories in Appendix Table~\ref{tab:gterr-full-appx}). Ground-truth correct here counts every correct answer, whereas $s^\ast$ uses the confident-correct share, so the percentages exceed $p_{\max}$ (for example 67.2\% against 50\% on VQA-RAD).

The error-class distributions show clear asymmetries under the original option order. On APTOS at $s^\ast$, non-ground-truth commitments concentrate on class C under both noise (61/97) and blur (21/22). On HAM10000, class A accounts for the largest share of errors under noise (109/220), while class C dominates under blur (115/225). On VQA-RAD, every non-ground-truth commitment under noise is class B (No, a Yes$\to$No flip), while 24/25 errors under blur are also class B, with the remaining error being class A (a No$\to$Yes flip). The option-order check below shows that this concentration does not persist under reversal.

\textbf{Post-Boundary Mirage Rate}

The \textbf{mirage rate} is the fraction of the baseline-correct population still committing (rather than abstaining) at the population-level breaking threshold $s^\ast$, regardless of correctness.

\vspace{-4pt}
\noindent
\begin{minipage}{\linewidth}
\centering
\tiny
\setlength{\tabcolsep}{3pt}
\renewcommand{\arraystretch}{0.75}
\captionof{table}{MedGemma: mirage rate at the breaking threshold $s^\ast$, with the share of wrong commitments and of confident wrong commitments ($H<\tau$) in the evaluated sample.}
\label{tab:mirage-rate}
\vspace{-2pt}
\begin{tabular}{llccccrr}
\toprule
\textbf{Pair} & \textbf{Modality} & $n$ & $s^\ast$ & \textbf{Commit / IDK} & \textbf{Mirage rate} & \textbf{Wrong} & \textbf{Conf. wrong} \\
\midrule
APTOS & Noise & 256 & 4 & 166 / 90 & \textbf{64.8\%} & 37.9\% & 16.8\% \\
APTOS & Blur & 256 & 7 & 41 / 215 & \textbf{16.0\%} & 8.6\% & 1.2\% \\
HAM10000 & Noise & 346 & 3 & 346 / 0 & \textbf{100.0\%} & 63.6\% & 9.0\% \\
HAM10000 & Blur & 346 & 5 & 346 / 0 & \textbf{100.0\%} & 65.0\% & 17.1\% \\
VQA-RAD & Noise & 128 & 8 & 108 / 20 & \textbf{84.4\%} & 17.2\% & 6.2\% \\
VQA-RAD & Blur & 128 & 7 & 111 / 17 & \textbf{86.7\%} & 19.5\% & 8.6\% \\
\bottomrule
\end{tabular}
\end{minipage}
\vspace{2pt}

HAM10000 has a 100\% mirage rate under both modalities because MedGemma essentially never abstains there. APTOS shows the largest modality contrast: under blur, 84.0\% of examples have abstained by $s^\ast=7$, yielding a mirage rate of 16.0\%, compared with 35.2\% abstaining by $s^\ast=4$ under noise, yielding a mirage rate of 64.8\%. VQA-RAD remains predominantly committing at its breaking threshold under both modalities, with commitment rates of 84.4\% under noise and 86.7\% under blur.

The rates depend on the reference used to place the boundary (Appendix Table~\ref{tab:nulls}). With the natural-correct share in place of the confident-correct share, the rates become 43.8, 16.0, 99.7, 98.3, 78.9, and 71.1\%. With $p_{\max}$ taken over the full baseline-correct pool, APTOS blur rises from 16.0 to 71.9\% and APTOS noise to 85.2\%, so the lower end of the range is specific to the sample-matched null, while HAM10000 stays at or above 98\%. Bootstrap resampling places $s^\ast$ within one step of the reported value in at least 99.4\% of resamples, and at the exact value in 68.2 to 100\% (Appendix Table~\ref{tab:boot}). Four noise draws with different random seeds place $s^\ast$ at the same value for each dataset (4, 3, and 8), with mirage rates of 62.9 to 68.0\% (APTOS), 100\% (HAM10000), and 82.0 to 85.2\% (VQA-RAD) (Appendix Table~\ref{tab:draws}). Varying $\tau$ between 0.05 and 0.30 leaves HAM10000 at 100\%, whereas the rate falls as $\tau$ rises for the other pairs: from 93.6 to 64.8\% for APTOS noise, from 92.6 to 16.0\% for APTOS blur, and from 98.7 to 84.4 and 86.7\% for VQA-RAD (Appendix Table~\ref{tab:tau}).

\looseness=-1 We also examine whether the model's own uncertainty separates wrong commitments from correct ones. At $s^\ast$, wrong answers given with $H<\tau$ make up 1.2 to 17.1\% of the sample (Table~\ref{tab:mirage-rate}). Median entropy is higher for wrong than for correct answers in most pairs (HAM10000 noise 0.72 against 0.39, VQA-RAD blur 0.36 against 0.11), but not on APTOS noise (0.44 against 0.50). Beyond $s^\ast$, 78.5\% of the wrong HAM10000 blur answers still have $H<\tau$ (Appendix Table~\ref{tab:entropy}). Stated uncertainty therefore flags some wrong commitments but misses others, so it is only a partial safeguard.

\textbf{Abstention Propensity and Option Order}

\looseness=-1 Differences in abstention propensity account for much of the variation across dataset-modality pairs: MedGemma abstains on more than half of the degraded APTOS outputs (mean 56.6 to 62.3\%), on 12.4 to 17.7\% of VQA-RAD outputs, and almost never on HAM10000, and the mirage rate, the abstention-ending trajectories, the blackout behavior, and the feasibility of the probe below follow this ordering (Appendix Table~\ref{tab:propensity}). We ran the image sweeps on the same examples with the answer options in reverse order, so that the abstention option is shown first, and with a rotated order that places it in the middle (Appendix Tables~\ref{tab:order} and~\ref{tab:draws}). Under the reversed order only 85.5\% (APTOS), 39.6\% (HAM10000), and 67.2\% (VQA-RAD) of the sampled examples are still confident-correct at $s=0$ (86, 36, and 75\% under the rotated order), so membership in the baseline-correct pool depends on the option order.

\vspace{2pt}

On APTOS the blackout abstention disappears under both alternative orders (100 to 0\%). At and beyond the original $s^\ast$, the reversed run returns the displayed letter that was modal in the original run for 99.1\% (noise) and 85.1\% (blur) of outputs but the same class for only 0.0\% and 13.9\%, whereas the rotated run returns the same letter for 0.4\% and 0.0\% and the same class for 34.4\% and 81.3\%. The APTOS abstention therefore follows neither a single displayed letter nor the class across orders, and it depends on where the abstention option is shown. On HAM10000 and VQA-RAD no letter persists under either order (same-letter share 0.1 to 11.1\%), and the same-class share is 0.0 to 5.3\% under reversal and 0.0\% (HAM10000) and 35.8 to 43.1\% (VQA-RAD) under rotation. The same 120 clean VQA-RAD items receive No for 65.8\% in the original order and Yes for 77.5\% in the reversed order, so the direction of the VQA-RAD default depends on the option order.

We also apply the image degradations after resizing to the model input resolution (896 pixels). This shifts $s^\ast$ by one to two steps and the mirage rate at each setting's own $s^\ast$ by $-16.0$ to $+8.6$ points across the four noise draws and the blur sweeps (Appendix Table~\ref{tab:resolution}). The spread of $s^\ast$ across datasets falls from 5 to 3 steps under noise and rises from 2 to 3 under blur, so comparisons of breaking severity across datasets should be read as approximate.

\textbf{PHI and Baseline Correctness}

\looseness=-1 Mean text-only PHI over the full MedGemma dataset populations, rather than the baseline-correct subpopulations, is 0.07 nats for APTOS, 2.08 for HAM10000, and 1.84 for VQA-RAD (Appendix~D). Baseline-correct examples have higher PHI than the rest on VQA-RAD (mean 2.43 against 1.00, AUROC for baseline correctness 0.677, 95\% bootstrap interval 0.646 to 0.706). On APTOS the association is inverse (AUROC 0.294, interval 0.276 to 0.311), and on HAM10000 it is 0.464 (0.455 to 0.473). For commit versus abstain at $s^\ast$, the AUROC of text-only PHI is 0.670 (noise) and 0.653 (blur) on VQA-RAD and 0.406 and 0.395 on APTOS. PHI is therefore positively associated with baseline correctness and with commitment at the boundary on VQA-RAD and inversely associated with both on APTOS. Ground-truth class is also associated with trajectory patterns, for example 34/64 No-answer against 3/64 Yes-answer VQA-RAD examples staying correct through noise blackout (Appendix~K).

\textbf{Commit-versus-Abstain Probe at the Breaking Threshold}

\looseness=-1 We probe whether commit-versus-abstain behavior at $s^\ast$ is linearly decodable from residual-stream activations. Labels are the model's own commit/abstain outputs at $s^\ast$.

\vspace{2pt}

All four probed conditions exceed both chance and the class-only baseline. Activations are extracted at four token positions and for the clean input, with the layer and regularization strength chosen inside the training folds (Appendix Table~\ref{tab:probe-nested}). The noise conditions are extracted under four noise draws, with activations and labels computed from the same noisy images, so label counts differ slightly across draws (Appendix Table~\ref{tab:probe-nested}). The nested AUROC at the last prompt token is 0.895 to 0.953 (APTOS noise), 0.994 (APTOS blur), 0.966 to 0.970 (VQA-RAD noise), and 0.896 (VQA-RAD blur). It is lower at the last image token (0.396 to 0.905) and question token (0.747 to 0.983) and 0.991 to 1.000 at the answer prefix, which indicates decodability where the letter is produced and not prediction ahead of time. Activations from the clean input predict the behavior at $s^\ast$ with nested AUROC 0.713 to 0.825 across conditions and draws, against class-only baselines of 0.550 to 0.630 (APTOS noise), 0.768 (APTOS blur), 0.581 to 0.626 (VQA-RAD noise), and 0.576 (VQA-RAD blur), so they carry information beyond the class for VQA-RAD and APTOS noise and only marginally for APTOS blur. We do not extend the probe to language degradation, because pooled IDK rates are at most 13.3\% and no population-level $s^\ast$ anchors it, since WordNet accuracy remains high on VQA-RAD (87 to 92\%) and is non-monotonic on HAM10000. The probe establishes representational separability and not a causal mechanism, and APTOS decodability may partly reflect how the abstention option is displayed, since APTOS abstention changes with option order.

\textbf{Language Degradation}

\looseness=-1 On VQA-RAD only the question text is mutated. WordNet substitution is mild (87.5\% natural accuracy at $r=0.75$), and character corruption is more damaging and non-monotonic (43.8\% at $r=0.50$, 53.1\% at $r=0.75$), yet IDK peaks at only 13.3\%, so the tested question perturbations move examples toward incorrect commitment more than toward abstention (Appendix~K). On HAM10000, where MedGemma never abstains, blackout accuracy is 18.5\% with no protection, and 73.4\% with the answer-choice descriptions protected, which indicates reliance on the descriptions. Mutating each example independently gives the same qualitative pattern, but accuracies differ from the shared-mutation values by up to 21 points, whereas four independent per-example draws differ from one another by at most 3.9 points (Appendix Table~\ref{tab:lang-v2}).

\textbf{Conclusion} 

Progressive evidence degradation reveals heterogeneous post-boundary behavior in MedGemma: APTOS is dominated by abstention, HAM10000 by non-abstaining commitment, and VQA-RAD by persistent correctness, recovery, and mirage. Trajectories and errors are also class-dependent. In the original option order, essentially all post-boundary errors at $s^\ast$ on VQA-RAD are Yes$\to$No flips, and this concentration does not persist when the option order is reversed. Mirage rates at $s^\ast$ range from 16.0\% to 100.0\%, with strong modality differences in APTOS (16.0\% blur vs.\ 64.8\% noise) and predominantly committing behavior in VQA-RAD (84.4\% noise, 86.7\% blur). These rates depend on the reference used to place the boundary and on the option order, and the APTOS abstention does not persist when the abstention option is moved. Entropy-based uncertainty separates some incorrect commitments from correct answers but misses others. PHI is associated with baseline correctness (AUROC 0.677) and with commitment at the boundary (AUROC 0.653 to 0.670) on VQA-RAD. In addition, residual-stream probes decode commit versus abstain at $s^\ast$ across all four conditions (nested AUROC 0.895 to 0.994 at the last prompt token), exceeding class-only baselines. Together, these results offer a baseline-anchored framework for validating medical foundation models under partial observation, revealing signals that accuracy curves obscure.

\textbf{Limitations and Recommendations}

The framework characterizes behavior but does not establish whether visual evidence still influences predictions or why the model switches to a particular answer or abstention response. Image ablation, matched image replacement, and activation intervention could test this. Moreover, synthetic noise and blur may not capture clinical degradation, and $s^\ast$ depends on $p_{\max}$ and the sampled population. The option-order results show dependence without a mechanism. In addition, free generation returned no parseable letter for several LLaVA-Med and Ovis pairs, probes are fit to the model's own output at the same severity and under the original option order only, and the main language results use one mutation per condition shared by all examples, with independent per-example mutations differing from the shared mutation by up to 21 points for a given attack and protection setting. Future work should therefore test natural degradation, alternative nulls and thresholds, and more option permutations. Closed-ended tasks, explicit abstention, and a language-only PHI limit extension to open-ended clinical generation and causal claims, and downstream analyses cover only MedGemma with small probe samples, so replication across VLMs and larger cohorts is needed. Code and results are at \repourl.

\bibliographystyle{plainnat}
\bibliography{biblitex}

@article{li2023llavamed,
  title     = {LLaVA-Med: Training a Large Language-and-Vision Assistant for Biomedicine in One Day},
  author    = {Li, Chunyuan and Wong, Cliff and Zhang, Sheng and Usuyama, Naoto and Liu, Haotian and Yang, Jianwei and Naumann, Tristan and Poon, Hoifung and Gao, Jianfeng},
  journal   = {arXiv preprint arXiv:2306.00890},
  year      = {2023}
}

@article{sellergren2025medgemma,
  title     = {MedGemma Technical Report},
  author    = {Sellergren, Andrew and Kazemzadeh, Sahar and Jaroensri, Tiam and Kiraly, Atilla and Traverse, Madeleine and Kohlberger, Timo and Xu, Shawn and Jamil, Fayaz and Hughes, C{\'\i}an and Lau, Charles and others},
  journal   = {arXiv preprint arXiv:2507.05201},
  year      = {2025}
}

@article{liu2026medqdeg,
  title     = {MedQ-Deg: A Multidimensional Benchmark for Evaluating MLLMs Across Medical Image Quality Degradations},
  author    = {Liu, Jiyao and Ning, Junzhi and Ma, Chenglong and Qu, Wanying and Shen, Jianghan and Luo, Siqi and Wei, Jinjie and Ye, Jin and Li, Pengze and Li, Tianbin and Lin, Jiashi and Shan, Hongming and Luo, Xinzhe and Liu, Xiaohong and Liu, Lihao and He, Junjun and Xu, Ningsheng},
  journal   = {arXiv preprint arXiv:2603.07769},
  year      = {2026}
}

@article{asadi2026mirage,
  title     = {MIRAGE: The Illusion of Visual Understanding},
  author    = {Asadi, Mohammad and O'Sullivan, Jack W. and Cao, Fang and Nedaee, Tahoura and Rajabalifardi, Kamyar and Li, Fei-Fei and Adeli, Ehsan and Ashley, Euan},
  journal   = {arXiv preprint arXiv:2603.21687},
  year      = {2026}
}

@article{benlevi2026mirageprobes,
  title     = {Mirage Probes: How Vision Models Fake Visual Understanding},
  author    = {Ben-Levi, Daniel and Goldfeder, Judah and Zhao, Weiliang and Lapid, Raz and LeVi, Amit and Roush, Allen G. and Shwartz-Ziv, Ravid and Lipson, Hod},
  journal   = {arXiv preprint arXiv:2606.13870},
  year      = {2026}
}

@article{medvigil2026broken,
  title     = {MedVIGIL: Evaluating Trustworthy Medical VLMs Under Broken Visual Evidence},
  author    = {Jiang, Hanqi and Chen, Junhao and Kang, Mingyu and Kwon, Hyeokjae and Pan, Yi and Chen, Lifeng and You, Weihang and Gong, Haozhen and Yan, Ruiyu and Lv, Jinglei and Zhao, Lin and Ren, Hui and Li, Quanzheng and Liu, Tianming and Li, Xiang},
  journal   = {arXiv preprint arXiv:2605.07919},
  year      = {2026}
}

@inproceedings{wang2022selfconsistency,
  title     = {Self-Consistency Improves Chain of Thought Reasoning in Language Models},
  author    = {Wang, Xuezhi and Wei, Jason and Schuurmans, Dale and Le, Quoc and Chi, Ed and Narang, Sharan and Chowdhery, Aakanksha and Zhou, Denny},
  booktitle = {International Conference on Learning Representations (ICLR)},
  year      = {2023}
}

@inproceedings{taubenfeld2025confidence,
  title     = {Confidence Improves Self-Consistency in {LLM}s},
  author    = {Taubenfeld, Amir and Sheffer, Tom and Ofek, Eran and Feder, Amir and Goldstein, Ariel and Gekhman, Zorik and Yona, Gal},
  booktitle = {Findings of the Association for Computational Linguistics: ACL 2025},
  pages     = {20090--20111},
  year      = {2025}
}

@inproceedings{srinivasan2024selectiveselectivepredictionreducing,
  title     = {Selective ``Selective Prediction'': Reducing Unnecessary Abstention in Vision-Language Reasoning},
  author    = {Srinivasan, Tejas and Hessel, Jack and Gupta, Tanmay and Lin, Bill Yuchen and Choi, Yejin and Thomason, Jesse and Chandu, Khyathi Raghavi},
  booktitle = {Findings of the Association for Computational Linguistics: ACL 2024},
  pages     = {12935--12948},
  year      = {2024}
}

@article{liu2025abstain,
  title     = {Do not Abstain! Identify and Solve the Uncertainty},
  author    = {Liu, Jingyu and Peng, Jingquan and Wu, Xiaopeng and Li, Xubin and Ge, Tiezheng and Zheng, Bo and Liu, Yong},
  journal   = {arXiv preprint arXiv:2506.00780},
  year      = {2025}
}

@article{aravindan2026secondguess,
  title     = {Second Guess: Detecting Uncertainty Through Abstention and Answer Stability in Small Language Models},
  author    = {Vaithinathan Aravindan, Ashwath and Kejriwal, Mayank},
  journal   = {arXiv preprint arXiv:2605.25394},
  year      = {2026}
}

@inproceedings{tong2024eyeswideshut,
  title     = {Eyes Wide Shut? Exploring the Visual Shortcomings of Multimodal {LLM}s},
  author    = {Tong, Shengbang and Liu, Zhuang and Zhai, Yuexiang and Ma, Yi and LeCun, Yann and Xie, Saining},
  booktitle = {Proceedings of the IEEE/CVF Conference on Computer Vision and Pattern Recognition (CVPR)},
  pages     = {9568--9578},
  year      = {2024}
}

@inproceedings{rubi2019,
  title     = {{RUB}i: Reducing Unimodal Biases for Visual Question Answering},
  author    = {Cad{\`e}ne, R{\'e}mi and Dancette, Corentin and Ben-younes, Hedi and Cord, Matthieu and Parikh, Devi},
  booktitle = {Advances in Neural Information Processing Systems (NeurIPS)},
  volume    = {32},
  pages     = {841--852},
  year      = {2019}
}

@inproceedings{si2022languagepriornotonly,
  title     = {Language Prior Is Not the Only Shortcut: A Benchmark for Shortcut Learning in {VQA}},
  author    = {Si, Qingyi and Meng, Fandong and Zheng, Mingyu and Lin, Zheng and Liu, Yuanxin and Fu, Peng and Cao, Yanan and Wang, Weiping and Zhou, Jie},
  booktitle = {Findings of the Association for Computational Linguistics: EMNLP 2022},
  pages     = {3698--3712},
  year      = {2022}
}

@article{radimagenetvqa2026,
  title     = {RadImageNet-{VQA}: A Large-Scale {CT} and {MRI} Dataset for Radiologic Visual Question Answering},
  author    = {Butsanets, L{\'e}o and Corbi{\`e}re, Charles and Khlaut, Julien and Manceron, Pierre and Dancette, Corentin},
  journal   = {Proceedings of Machine Learning Research},
  volume    = {315},
  pages     = {3036--3068},
  year      = {2026},
  note      = {arXiv:2512.17396}
}

@inproceedings{medbiasx2025,
  title     = {Med-{B}ias{X}: Robust Medical Visual Question Answering with Language Biases},
  author    = {Zhu, Huanjia and Liu, Yishu and Zhou, Chengju and Lu, Guangming and Chen, Bingzhi},
  booktitle = {Medical Image Computing and Computer Assisted Intervention -- MICCAI 2025},
  volume    = {LNCS 15973},
  pages     = {369--378},
  publisher = {Springer, Cham},
  year      = {2025},
  doi       = {10.1007/978-3-032-05185-1_36}
}

@inproceedings{decoct2025,
  title     = {Eliminating Language Bias for Medical Visual Question Answering with Counterfactual Contrastive Training},
  author    = {Wan, Xingyu and Teng, Qiaoying and Chen, Jun and Lu, Yonghan and Yuan, Deqi and Liu, Zhe},
  booktitle = {Medical Image Computing and Computer Assisted Intervention -- MICCAI 2025},
  volume    = {LNCS 15965},
  pages     = {194--204},
  publisher = {Springer Nature Switzerland},
  year      = {2025}
}

@article{morenocencerrado2025noanswer,
  title     = {No Answer Needed: Predicting {LLM} Answer Accuracy from Question-Only Linear Probes},
  author    = {Moreno Cencerrado, Iv{\'a}n Vicente and Padr{\'e}s Masdemont, Arnau and Gonzalvez Hawthorne, Anton and Africa, David Demitri and Pacchiardi, Lorenzo},
  journal   = {arXiv preprint arXiv:2509.10625},
  year      = {2025}
}

@article{wang2025vision,
  title={Are vision language models robust to uncertain inputs?},
  author={Wang, Xi and Nalisnick, Eric},
  journal={arXiv preprint arXiv:2505.11804},
  year={2025},
  doi={10.48550/arXiv.2505.11804}
}

@article{lau2018vqarad,
  title     = {A dataset of clinically generated visual questions and answers about radiology images},
  author    = {Lau, Jason J. and Gayen, Soumya and Ben Abacha, Asma and Demner-Fushman, Dina},
  journal   = {Scientific Data},
  volume    = {5},
  pages     = {180251},
  year      = {2018},
  doi       = {10.1038/sdata.2018.251}
}

@article{tschandl2018ham10000,
  title     = {The {HAM10000} dataset, a large collection of multi-source dermatoscopic images of common pigmented skin lesions},
  author    = {Tschandl, Philipp and Rosendahl, Cliff and Kittler, Harald},
  journal   = {Scientific Data},
  volume    = {5},
  pages     = {180161},
  year      = {2018},
  doi       = {10.1038/sdata.2018.161}
}

@misc{aptos2019,
  title        = {{APTOS} 2019 Blindness Detection},
  author       = {{Asia Pacific Tele-Ophthalmology Society}},
  year         = {2019},
  howpublished = {Kaggle competition},
  url          = {https://www.kaggle.com/competitions/aptos2019-blindness-detection}
}

@article{lu2025ovis25,
  title     = {{Ovis2.5} Technical Report},
  author    = {Lu, Shiyin and others},
  journal   = {arXiv preprint arXiv:2508.11737},
  year      = {2025}
}

\textbf{Appendix}

\textbf{Appendix A: Trajectory Classification and IDK Encoding}

Trajectory categories follow the classification procedure specified in the Methodology. Low-confidence correct predictions ($\max_k p_k<0.5$, $\hat y=$ answer) are assigned to IDK, while low-confidence wrong predictions are assigned to wrong. The trajectory counts and the typology figures use this encoding. The class-level IDK heatmaps, the mirage-rate calculations, and the probe labels use the plain abstention flag and do not use the $0.5$ maximum-probability criterion. The four entropy-based response regimes below are therefore distinct from this three-state trajectory encoding. Finally, the evaluated checkpoints are MedGemma 1.5 4B-it, LLaVA-Med v1.5 (Mistral-7B, Hugging Face conversion), and Ovis2.5-2B.

As severity increases, individual examples can occupy different response regimes at each severity level. We distinguish four regimes: (1) \textbf{High-entropy response.} The model selects a substantive answer, but the answer distribution fails to concentrate on a single choice ($H \geq \tau$). (2) \textbf{Incorrect commitment.} The model selects a definite incorrect answer with concentrated probability ($H < \tau$). (3) \textbf{Correct.} The model selects the correct answer with concentrated probability ($H < \tau$). (4) \textbf{Abstention (IDK).} The model produces an abstention response. These regimes describe response behavior at individual severity levels, and Table~\ref{tab:regimes} reports their shares at $s^\ast$. The trajectory types defined in the Methodology describe sequences of the three-state trajectory encoding across severity levels.

\vspace{4pt}
\noindent
\begin{minipage}{\linewidth}
\centering
\tiny
\setlength{\tabcolsep}{3pt}
\renewcommand{\arraystretch}{0.75}
\captionof{table}{MedGemma: share of the evaluated sample in each response regime at $s^\ast$.}
\label{tab:regimes}
\vspace{-2pt}
\begin{tabular}{llcccccc}
\toprule
\textbf{Pair} & \textbf{Modality} & $n$ & $s^\ast$ & \textbf{Abstention} & \textbf{High-entropy} & \textbf{Incorrect commit.} & \textbf{Correct} \\
\midrule
APTOS & Noise & 256 & 4 & 35.2\% & 38.3\% & 16.8\% & 9.8\% \\
APTOS & Blur & 256 & 7 & 84.0\% & 10.2\% & 1.2\% & 4.7\% \\
HAM10000 & Noise & 346 & 3 & 0.0\% & 76.6\% & 9.0\% & 14.5\% \\
HAM10000 & Blur & 346 & 5 & 0.0\% & 66.8\% & 17.1\% & 16.2\% \\
VQA-RAD & Noise & 128 & 8 & 15.6\% & 37.5\% & 6.2\% & 40.6\% \\
VQA-RAD & Blur & 128 & 7 & 13.3\% & 29.7\% & 8.6\% & 48.4\% \\
\bottomrule
\end{tabular}
\end{minipage}

\textbf{Trajectory type definitions.} The six types, applied in this precedence, are: (1) \textbf{Persistent correct:} Correct at every severity level, including blackout. (2) \textbf{Direct mirage:} Departs from correct into the wrong state at some severity and remains wrong through blackout, never subsequently abstaining. (3) \textbf{Correct recovery:} Departs from correct into wrong or IDK-like states at an intermediate severity, then returns to the correct state at a later severity, including near-blackout recovery. This trajectory is interpreted descriptively, and a recovery is not by itself evidence that the model used language priors. (4) \textbf{Mirage dissolves:} Departs from correct into a wrong commitment, but eventually transitions into a stable IDK/abstain state rather than remaining wrong or recovering. (5) \textbf{Consistent abstention:} Departs from correct directly into a stable IDK/abstain state, without ever passing through a wrong commitment. (6) \textbf{Oscillatory:} Repeatedly transitions among multiple non-correct states without fitting any of the preceding trajectory types. Because the blackout image is identical for every example, types (1) to (4) depend on a terminal state that is a function of the prompt and class only, so Appendix~G also reports the counts with the blackout step removed.

\noindent
\begin{minipage}{\linewidth}
\centering
\tiny
\renewcommand{\arraystretch}{0.75}
\setlength{\tabcolsep}{2pt}
\captionof{table}{MedGemma: per-example breaking severity, the first severity at which an example is no longer natural-correct (answered and correct). Median, IQR, and number of examples that never break, including at blackout.}
\label{tab:boundary-summary}
\vspace{-2pt}
\begin{tabular}{llccc}
\toprule
\textbf{Pair} & \textbf{Modality} & \textbf{Median severity} & \textbf{IQR} & \textbf{Never breaks} \\
\midrule
APTOS & Noise & 4 & [3, 5] & 0 \\
APTOS & Blur & 6 & [5, 7] & 0 \\
HAM10000 & Noise & 2 & [2, 4] & 0 \\
HAM10000 & Blur & 5 & [4, 6] & 0 \\
VQA-RAD & Noise & 9 & [7, 10] & 37/128 \\
VQA-RAD & Blur & 8 & [7, 9] & 32/128 \\
\bottomrule
\end{tabular}
\end{minipage}

\textbf{Appendix B: Post-Boundary Commitments and Selected-Severity Trajectories}

\vspace{-4pt}
\noindent
\begin{minipage}{\linewidth}
\centering
\tiny
\setlength{\tabcolsep}{2pt}
\renewcommand{\arraystretch}{0.75}
\captionof{table}{MedGemma: ground-truth / non-ground-truth / abstention split at each pair's population-level breaking threshold $s^\ast$, with the class distribution of non-ground-truth commitments. Letters follow the option mapping in the Methodology.}
\label{tab:gt-error-sstar}
\vspace{-2pt}
\begin{tabular}{llcccccl}
\toprule
\textbf{Pair} & \textbf{Modality} & $s^\ast$ & \textbf{GT\%} & \textbf{Non-GT\%} & \textbf{Abstain\%} & \textbf{Error class dist.} \\
\midrule
APTOS & Noise & 4 & 27.0 & 37.9 & 35.2 & A: 36, C: 61 \\
APTOS & Blur & 7 & 7.4 & 8.6 & 84.0 & C: 21, E: 1 \\
HAM10000 & Noise & 3 & 36.4 & 63.6 & 0.0 & C: 60, A: 109, E: 44, F: 1, B: 6 \\
HAM10000 & Blur & 5 & 35.0 & 65.0 & 0.0 & C: 115, A: 86, F: 5, B: 7, E: 12 \\
VQA-RAD & Noise & 8 & 67.2 & 17.2 & 15.6 & B: 22 (all) \\
VQA-RAD & Blur & 7 & 67.2 & 19.5 & 13.3 & B: 24, A: 1 \\
\bottomrule
\end{tabular}
\end{minipage}
\vspace{2pt}

This appendix also provides the ground-truth (GT), non-ground-truth (nGT), and abstention (Ab) decomposition at selected severities beyond $s^\ast$ (Table~\ref{tab:gterr-full-appx}).

\vspace{-4pt}
\noindent
\begin{minipage}{\linewidth}
\centering
\tiny
\setlength{\tabcolsep}{1pt}
\renewcommand{\arraystretch}{0.65}
\captionof{table}{MedGemma: ground-truth (GT), non-ground-truth (nGT), and abstain (Ab) percentage at selected severities, with BO denoting blackout.}
\label{tab:gterr-full-appx}
\vspace{-2pt}
\begin{tabular}{ll ccc ccc ccc ccc ccc}
\toprule
& & \multicolumn{3}{c}{s=0} & \multicolumn{3}{c}{s=4} & \multicolumn{3}{c}{s=7} & \multicolumn{3}{c}{s=10} & \multicolumn{3}{c}{BO} \\
\textbf{Pair} & \textbf{Mod.} & GT & nGT & Ab & GT & nGT & Ab & GT & nGT & Ab & GT & nGT & Ab & GT & nGT & Ab \\
\midrule
APTOS & Noise & 100.0 & 0.0 & 0.0 & \textbf{27.0} & \textbf{37.9} & \textbf{35.2} & 5.5 & 17.6 & 77.0 & 3.9 & 7.0 & 89.1 & 0.0 & 0.0 & 100.0 \\
APTOS & Blur & 100.0 & 0.0 & 0.0 & 93.0 & 2.0 & 5.1 & \textbf{7.4} & \textbf{8.6} & \textbf{84.0} & 0.0 & 2.3 & 97.7 & 0.0 & 0.0 & 100.0 \\
HAM10000 & Noise & 100.0 & 0.0 & 0.0 & 28.9 & 71.1 & 0.0 & 23.1 & 76.6 & 0.3 & 18.2 & 81.5 & 0.3 & 18.5 & 81.5 & 0.0 \\
HAM10000 & Blur & 100.0 & 0.0 & 0.0 & 72.0 & 28.0 & 0.0 & 18.8 & 79.8 & 1.4 & 17.1 & 80.9 & 2.0 & 18.5 & 81.5 & 0.0 \\
VQA-RAD & Noise & 100.0 & 0.0 & 0.0 & 94.5 & 5.5 & 0.0 & 76.6 & 17.2 & 6.2 & 48.4 & 30.5 & 21.1 & 63.3 & 25.0 & 11.7 \\
VQA-RAD & Blur & 100.0 & 0.0 & 0.0 & 100.0 & 0.0 & 0.0 & \textbf{67.2} & \textbf{19.5} & \textbf{13.3} & 32.8 & 24.2 & 43.0 & 63.3 & 25.0 & 11.7 \\
\bottomrule
\end{tabular}
\end{minipage}
{\tiny\noindent HAM10000 Noise has $s^\ast=3$, HAM10000 Blur has $s^\ast=5$, and VQA-RAD Noise has $s^\ast=8$, which fall between the selected columns shown above. Exact GT/nGT/Ab values at $s^\ast$ for every pair are in Table~\ref{tab:gt-error-sstar}.}

\textbf{Appendix C: Natural vs.\ Forced-Commit Behavior at Blackout}

This appendix compares naturally generated behavior, in which IDK/abstention is permitted, with forced-commit behavior at full blackout. "Same?" indicates whether natural and forced-commit accuracy coincide.

\vspace{-4pt}
\noindent
\begin{minipage}{\linewidth}
\centering
\tiny
\setlength{\tabcolsep}{2pt}
\renewcommand{\arraystretch}{0.75}
\captionof{table}{MedGemma: natural (IDK-permitted) versus forced-commit behavior at full blackout, for all reported pairs. Forced commitment here restricts the softmax of the natural prompt to the substantive options.}
\label{tab:idk-vs-noidk-appx}
\vspace{-2pt}
\begin{tabular}{llccc}
\toprule
\textbf{Pair} & \textbf{Channel} & \textbf{Natural (IDK-permitted)} & \textbf{Forced-commit} & \textbf{Same?} \\
\midrule
APTOS & Image (blackout) & 100\% abstain & 25.0\% acc., canned `C' & No \\
APTOS & Language (\texttt{protect\_none}, BO) & 3.9\% IDK, 24.6\% acc. & 24.6\% acc. & Yes \\
APTOS & Language (\texttt{protect\_idk}, BO) & 41.0\% IDK, 17.6\% acc. & 24.6\% acc. & No \\
HAM10000 & Image (blackout) & 0\% abstain, 18.5\% acc. & 18.5\% acc., canned `A' & Yes \\
HAM10000 & Language (\texttt{protect\_none}, BO) & 0.0\% IDK, 18.5\% acc. & 18.5\% acc. & Yes \\
HAM10000 & Language (\texttt{protect\_idk}, BO) & 0.0\% IDK, 18.2\% acc. & 18.5\% acc. & Yes$^\ast$ \\
HAM10000 & Language (\texttt{protect\_desc}, BO) & 0.0\% IDK, 73.4\% acc. & 73.4\% acc. & Yes \\
HAM10000 & Language (\texttt{protect\_all}, BO) & 0.0\% IDK, 98.8\% acc. & 98.8\% acc. & Yes \\
VQA-RAD & Image (blackout) & 11.7\% abstain, 63.3\% acc. & 71.9\% acc., split A/B & No \\
VQA-RAD & Language (blackout, pooled) & 3.1\% IDK, 50.8\% acc. & 53.1\% acc. & No \\
\bottomrule
\end{tabular}
\end{minipage}
{\tiny\noindent $^\ast$The 0.3 percentage-point difference is within rounding tolerance.}

\textbf{Appendix D: PHI Means Across Models and Full-Dataset Populations}

Table~\ref{tab:phi-all-models-appx} is the image-present variant, computed on the full per-dataset populations (APTOS $n=2930$, HAM10000 $n=10015$, VQA-RAD $n=1191$) with the clean image present in both prompts. The text-only variant used in the main text, available for MedGemma only, has means of 0.073 nats (APTOS), 2.077 (HAM10000), and 1.840 (VQA-RAD), with means for baseline-correct and other examples of $-0.022$ and 0.192 (APTOS), 1.911 and 2.159 (HAM10000), and 2.427 and 0.997 (VQA-RAD). On the capped examples the two variants correlate weakly (Pearson $r=-0.25$ APTOS, $0.49$ HAM10000, $-0.08$ VQA-RAD), and image-present PHI separates stay-correct from post-boundary-wrong examples on VQA-RAD with AUROC 0.692 (noise) and 0.739 (blur).

\vspace{-4pt}
\noindent
\begin{minipage}{\linewidth}
\centering
\tiny
\setlength{\tabcolsep}{4pt}
\renewcommand{\arraystretch}{0.75}
\captionof{table}{Mean image-present PHI (nats) by model and dataset, computed over the full dataset population.}
\label{tab:phi-all-models-appx}
\vspace{-2pt}
\begin{tabular}{llcc}
\toprule
\textbf{Model} & \textbf{Dataset} & $n$ & \textbf{Mean PHI} \\
\midrule
MedGemma & APTOS & 2930 & 0.316 \\
MedGemma & HAM10000 & 10015 & 1.456 \\
MedGemma & VQA-RAD & 1191 & 2.546 \\
LLaVA-Med & APTOS & 2930 & 0.046 \\
LLaVA-Med & HAM10000 & 10015 & -0.955 \\
LLaVA-Med & VQA-RAD & 1191 & -0.258 \\
Ovis & APTOS & 2930 & -1.809 \\
Ovis & HAM10000 & 10015 & 0.530 \\
Ovis & VQA-RAD & 1191 & -2.368 \\
\bottomrule
\end{tabular}
\end{minipage}

\textbf{Appendix E: APTOS and HAM10000 Language Degradation}

\vspace{-4pt}
\noindent
\begin{minipage}{\linewidth}
\centering
\tiny
\setlength{\tabcolsep}{2pt}
\renewcommand{\arraystretch}{0.75}
\captionof{table}{MedGemma: APTOS text degradation, full $n=256$ population, natural (IDK-permitted) versus forced-commit accuracy, aggregated across the four protection configurations.}
\label{tab:aptos-lang-full-appx}
\vspace{-2pt}
\begin{tabular}{lccc}
\toprule
\textbf{Attack (severity)} & \textbf{IDK\%} & \textbf{Natural acc.} & \textbf{Forced acc.} \\
\midrule
wordnet (0.25) & 0.0 & 94.7\% & 94.7\% \\
wordnet (0.50) & 0.0 & 88.6\% & 88.6\% \\
wordnet (0.75) & 0.0 & 80.5\% & 80.5\% \\
char (0.25) & 0.2 & 82.9\% & 83.1\% \\
char (0.50) & 0.0 & 58.1\% & 58.1\% \\
char (0.75) & 10.5 & 45.0\% & 49.0\% \\
blackout (1.00) & 11.2 & 55.7\% & 57.4\% \\
\bottomrule
\end{tabular}
\end{minipage}
\vspace{4pt}

\noindent
\begin{minipage}{\linewidth}
\centering
\tiny
\setlength{\tabcolsep}{2pt}
\renewcommand{\arraystretch}{0.75}
\captionof{table}{MedGemma: HAM10000 text degradation, full $n=346$ population, natural (IDK-permitted) versus forced-commit accuracy, aggregated across the four protection configurations.}
\label{tab:ham-lang-full-appx}
\vspace{-2pt}
\begin{tabular}{lccc}
\toprule
\textbf{Attack (severity)} & \textbf{IDK\%} & \textbf{Natural acc.} & \textbf{Forced acc.} \\
\midrule
wordnet (0.25) & 0.0 & 80.7\% & 80.7\% \\
wordnet (0.50) & 0.0 & 91.3\% & 91.3\% \\
wordnet (0.75) & 0.0 & 70.8\% & 70.8\% \\
char (0.25) & 0.0 & 79.6\% & 79.6\% \\
char (0.50) & 0.0 & 58.3\% & 58.3\% \\
char (0.75) & 0.0 & 56.3\% & 56.4\% \\
blackout (1.00) & 0.0 & 52.2\% & 52.3\% \\
\bottomrule
\end{tabular}
\end{minipage}

\textbf{Appendix F: Robustness of the Breaking Threshold}\label{app:sstar}

Table~\ref{tab:nulls} reports $s^\ast$ and the mirage rate under two statistics (confident-correct, used in the main text, and natural-correct) and two references ($p_{\max}$ of the evaluated sample and of the full baseline-correct pool). Table~\ref{tab:boot} gives bootstrap intervals for $s^\ast$ from 2000 class-stratified resamples of the examples. Table~\ref{tab:tau} shrinks the pool to examples with clean-input entropy below $\tau'\leq 0.30$, which is the only direction available without new sweeps.

\vspace{-4pt}
\noindent
\begin{minipage}{\linewidth}
\centering
\tiny
\setlength{\tabcolsep}{2pt}
\renewcommand{\arraystretch}{0.75}
\captionof{table}{MedGemma: $s^\ast$ and mirage rate (\%) under alternative definitions. The first block uses $p_{\max}$ of the evaluated sample, the second block $p_{\max}$ of the full baseline-correct pool. Conf. is confident-correct, Nat. is natural-correct.}
\label{tab:nulls}
\vspace{-2pt}
\begin{tabular}{llc cc cc c cc cc}
\toprule
& & \multicolumn{5}{c}{Sample null} & \multicolumn{5}{c}{Pool null} \\
\textbf{Pair} & \textbf{Mod.} & $p_{\max}$ & Conf. $s^\ast$ & Mirage & Nat. $s^\ast$ & Mirage & $p_{\max}$ & Conf. $s^\ast$ & Mirage & Nat. $s^\ast$ & Mirage \\
\midrule
APTOS & Noise & 0.250 & 4 & 64.8 & 5 & 43.8 & 0.678 & 3 & 85.2 & 3 & 85.2 \\
APTOS & Blur & 0.250 & 7 & 16.0 & 7 & 16.0 & 0.678 & 5 & 71.9 & 5 & 71.9 \\
HAM10000 & Noise & 0.185 & 3 & 100.0 & 10 & 99.7 & 0.758 & 1 & 100.0 & 2 & 100.0 \\
HAM10000 & Blur & 0.185 & 5 & 100.0 & 8 & 98.3 & 0.758 & 3 & 100.0 & 4 & 100.0 \\
VQA-RAD & Noise & 0.500 & 8 & 84.4 & 10 & 78.9 & 0.628 & 7 & 93.8 & 9 & 80.5 \\
VQA-RAD & Blur & 0.500 & 7 & 86.7 & 8 & 71.1 & 0.628 & 7 & 86.7 & 8 & 71.1 \\
\bottomrule
\end{tabular}
\end{minipage}
\vspace{4pt}

\noindent
\begin{minipage}{\linewidth}
\centering
\tiny
\setlength{\tabcolsep}{3pt}
\renewcommand{\arraystretch}{0.75}
\captionof{table}{MedGemma: bootstrap 95\% interval of $s^\ast$, share of resamples at the reported value, and share within one severity step.}
\label{tab:boot}
\vspace{-2pt}
\begin{tabular}{llcccc}
\toprule
\textbf{Pair} & \textbf{Modality} & $s^\ast$ & \textbf{95\% interval} & \textbf{At $s^\ast$ (\%)} & \textbf{Within $\pm1$ (\%)} \\
\midrule
APTOS & Noise & 4 & [4, 4] & 100.0 & 100.0 \\
APTOS & Blur & 7 & [6, 7] & 76.8 & 100.0 \\
HAM10000 & Noise & 3 & [3, 3] & 99.2 & 100.0 \\
HAM10000 & Blur & 5 & [5, 6] & 94.5 & 99.4 \\
VQA-RAD & Noise & 8 & [7, 8] & 91.5 & 100.0 \\
VQA-RAD & Blur & 7 & [7, 8] & 68.2 & 100.0 \\
\bottomrule
\end{tabular}
\end{minipage}
\vspace{4pt}

\noindent
\begin{minipage}{\linewidth}
\centering
\tiny
\setlength{\tabcolsep}{3pt}
\renewcommand{\arraystretch}{0.75}
\captionof{table}{MedGemma: entropy-threshold sensitivity on the shrinking pool. Each entry is pool size / $s^\ast$ / mirage rate (\%) for $\tau'=0.05, 0.10, 0.20, 0.30$.}
\label{tab:tau}
\vspace{-2pt}
\begin{tabular}{llcccc}
\toprule
\textbf{Pair} & \textbf{Mod.} & $\tau'=0.05$ & $\tau'=0.10$ & $\tau'=0.20$ & $\tau'=0.30$ \\
\midrule
APTOS & Noise & 94 / 2 / 93.6 & 178 / 3 / 85.4 & 230 / 4 / 66.5 & 256 / 4 / 64.8 \\
APTOS & Blur & 94 / 5 / 92.6 & 178 / 6 / 53.9 & 230 / 6 / 51.7 & 256 / 7 / 16.0 \\
HAM10000 & Noise & 50 / 2 / 100.0 & 159 / 3 / 100.0 & 277 / 3 / 100.0 & 346 / 3 / 100.0 \\
HAM10000 & Blur & 50 / 4 / 100.0 & 159 / 5 / 100.0 & 277 / 5 / 100.0 & 346 / 5 / 100.0 \\
VQA-RAD & Noise & 76 / 5 / 98.7 & 104 / 6 / 97.1 & 121 / 7 / 93.4 & 128 / 8 / 84.4 \\
VQA-RAD & Blur & 76 / 6 / 98.7 & 104 / 7 / 88.5 & 121 / 7 / 87.6 & 128 / 7 / 86.7 \\
\bottomrule
\end{tabular}
\end{minipage}

\textbf{Appendix G: Trajectories Without the Blackout Step and Under a Symmetric Encoding}\label{app:traj}

\vspace{-4pt}
\noindent
\begin{minipage}{\linewidth}
\centering
\tiny
\setlength{\tabcolsep}{2pt}
\renewcommand{\arraystretch}{0.75}
\captionof{table}{MedGemma: trajectory breakdown by pair and modality, with the blackout step included.}
\label{tab:trajectory-breakdown}
\vspace{-2pt}
\begin{tabular}{llcccccc}
\toprule
\textbf{Pair} & \textbf{Modality} & \shortstack{\textbf{Pers.}\\ \textbf{corr.}} & \shortstack{\textbf{Direct}\\ \textbf{mir.}} & \shortstack{\textbf{Correct}\\ \textbf{rec.}} & \shortstack{\textbf{Mir.}\\ \textbf{diss.}} & \shortstack{\textbf{Cons.}\\ \textbf{abst.}} & \textbf{Osc.} \\
\midrule
APTOS & Noise & 0 & 0 & 0 & 158 & 86 & 12 \\
APTOS & Blur & 0 & 0 & 0 & 74 & 133 & 49 \\
HAM10000 & Noise & 0 & 204 & 64 & 0 & 0 & 78 \\
HAM10000 & Blur & 0 & 233 & 64 & 0 & 0 & 49 \\
VQA-RAD & Noise & 37 & 27 & 42 & 5 & 11 & 6 \\
VQA-RAD & Blur & 32 & 20 & 47 & 4 & 13 & 12 \\
\bottomrule
\end{tabular}
\end{minipage}

Table~\ref{tab:traj-robust} gives the six trajectory counts for severities $s=0,\dots,13$ only (the final state is then the state at $s=13$), and under a symmetric encoding in which low-confidence wrong answers are also treated as IDK. Under the symmetric encoding the APTOS and VQA-RAD counts barely change, whereas HAM10000 direct mirage falls (204 to 127 under noise) and oscillatory rises (78 to 155), so the HAM10000 split between these two types depends on the encoding. Without the blackout step the HAM10000 recoveries (50 under noise, 29 under blur) are all class C examples, in contrast to the class A recoveries with blackout. The state at $s=13$ is itself a near-constant output. The blackout state of each pair is stated in the main text.

\vspace{-4pt}
\noindent
\begin{minipage}{\linewidth}
\centering
\tiny
\setlength{\tabcolsep}{2pt}
\renewcommand{\arraystretch}{0.75}
\captionof{table}{MedGemma: trajectory counts excluding the blackout step and under the symmetric encoding. Columns follow Appendix Table~\ref{tab:trajectory-breakdown}.}
\label{tab:traj-robust}
\vspace{-2pt}
\begin{tabular}{lllcccccc}
\toprule
\textbf{Pair} & \textbf{Mod.} & \textbf{Variant} & \shortstack{\textbf{Pers.}\\ \textbf{corr.}} & \shortstack{\textbf{Direct}\\ \textbf{mir.}} & \shortstack{\textbf{Correct}\\ \textbf{rec.}} & \shortstack{\textbf{Mir.}\\ \textbf{diss.}} & \shortstack{\textbf{Cons.}\\ \textbf{abst.}} & \textbf{Osc.} \\
\midrule
APTOS & Noise & no blackout & 0 & 0 & 0 & 158 & 86 & 12 \\
 & & symmetric & 0 & 0 & 0 & 151 & 93 & 12 \\
APTOS & Blur & no blackout & 0 & 11 & 43 & 62 & 133 & 7 \\
 & & symmetric & 0 & 0 & 0 & 71 & 136 & 49 \\
HAM10000 & Noise & no blackout & 11 & 251 & 50 & 0 & 0 & 34 \\
 & & symmetric & 0 & 127 & 64 & 0 & 0 & 155 \\
HAM10000 & Blur & no blackout & 32 & 258 & 29 & 0 & 0 & 27 \\
 & & symmetric & 0 & 217 & 64 & 0 & 0 & 65 \\
VQA-RAD & Noise & no blackout & 39 & 40 & 2 & 11 & 28 & 8 \\
 & & symmetric & 37 & 28 & 42 & 5 & 11 & 5 \\
VQA-RAD & Blur & no blackout & 34 & 27 & 20 & 9 & 26 & 12 \\
 & & symmetric & 32 & 20 & 47 & 4 & 13 & 12 \\
\bottomrule
\end{tabular}
\end{minipage}

\textbf{Appendix H: Entropy of Committed Answers}

\vspace{-4pt}
\noindent
\begin{minipage}{\linewidth}
\centering
\tiny
\setlength{\tabcolsep}{2pt}
\renewcommand{\arraystretch}{0.75}
\captionof{table}{MedGemma: median entropy in nats of committed (non-abstaining) answers and, in parentheses, the share with $H<\tau$. At $s^\ast$ and over $s^\ast,\dots,13$, split by correct and wrong.}
\label{tab:entropy}
\vspace{-2pt}
\begin{tabular}{llcccc}
\toprule
& & \multicolumn{2}{c}{At $s^\ast$} & \multicolumn{2}{c}{Over $s^\ast$ to 13} \\
\textbf{Pair} & \textbf{Mod.} & Correct & Wrong & Correct & Wrong \\
\midrule
APTOS & Noise & 0.50 (36.2\%) & 0.44 (44.3\%) & 0.57 (23.5\%) & 0.68 (21.3\%) \\
APTOS & Blur & 0.15 (63.2\%) & 0.49 (13.6\%) & 0.61 (19.0\%) & 0.45 (33.9\%) \\
HAM10000 & Noise & 0.39 (39.7\%) & 0.72 (14.1\%) & 0.25 (59.8\%) & 0.33 (46.4\%) \\
HAM10000 & Blur & 0.35 (46.3\%) & 0.59 (26.2\%) & 0.10 (81.8\%) & 0.10 (78.5\%) \\
VQA-RAD & Noise & 0.24 (60.5\%) & 0.41 (36.4\%) & 0.23 (58.2\%) & 0.40 (36.9\%) \\
VQA-RAD & Blur & 0.11 (72.1\%) & 0.36 (44.0\%) & 0.19 (63.6\%) & 0.25 (56.8\%) \\
\bottomrule
\end{tabular}
\end{minipage}

\textbf{Appendix I: Probe Layer-Wise Results}

Figure~\ref{fig:probe-layers} shows the cross-validated AUROC at every decoder layer for the four probed conditions. For the noise conditions the curves use the last prompt token and four noise draws, with activations and labels from the same noisy images. On APTOS noise 16 to 24 of 34 layers exceed 0.9 (best layer 0.926 to 0.963), and on VQA-RAD noise 19 of 34 do (best layer 0.976 to 0.977). For blur, 21 of 34 layers exceed 0.9 on APTOS and none does on VQA-RAD (best layer 0.878).

\noindent
\begin{minipage}{\linewidth}
\centering
\includegraphics[width=0.6\linewidth]{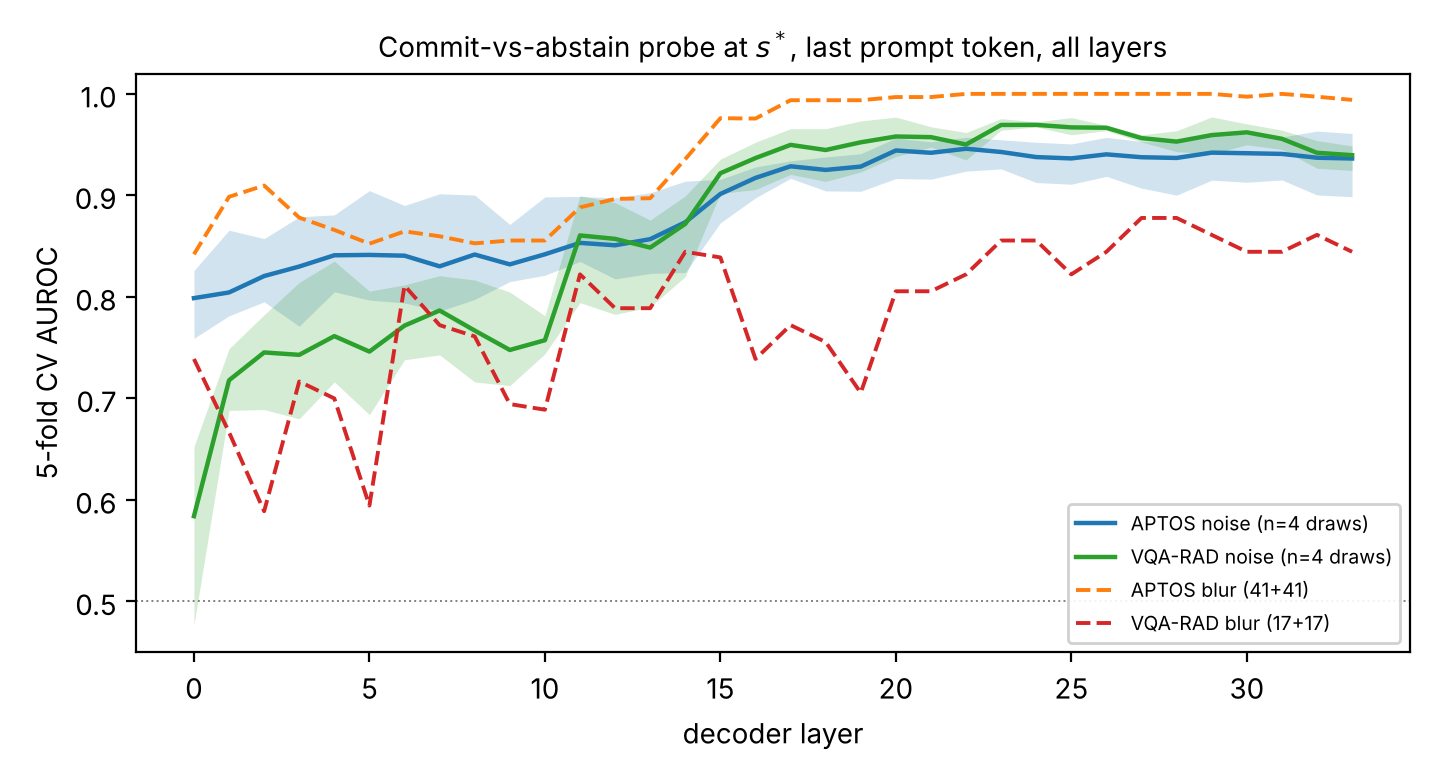}
\captionof{figure}{MedGemma: 5-fold cross-validated AUROC of the commit-vs-abstain probe at each decoder layer, at $s^\ast$. Noise curves show the mean and the range over four draws (seeds 0 to 3), and blur curves (dashed) come from class-balanced subsamples of 41+41 (APTOS) and 17+17 (VQA-RAD) examples. Layer 0 gives .759 to .826 (APTOS noise), .477 to .652 (VQA-RAD noise), 0.842 (APTOS blur), and 0.739 (VQA-RAD blur).}
\label{fig:probe-layers}
\end{minipage}

\needspace{8\baselineskip}\textbf{Appendix J: Option-Order, Resolution, Probe, and Language Runs}

\noindent
\begin{minipage}{\linewidth}
\centering
\tiny
\setlength{\tabcolsep}{2pt}
\renewcommand{\arraystretch}{0.75}
\captionof{table}{MedGemma: image sweeps under the original and the reversed option order. Mirage rates are at the original $s^\ast$, and the blackout column gives the percent abstaining at $s=14$. Same class and same letter give the share of outputs at and beyond the original $s^\ast$ in which the reversed run returns the class, or the displayed letter, that was modal in the original run. Noise rows use the original noise draw. Conf.-correct at $s=0$ gives the percent of the sampled examples that are confident-correct (answered, $H<\tau$, correct) at $s=0$ under the reversed order, against 100\% under the original order by construction.}
\label{tab:order}
\vspace{-2pt}
\begin{tabular}{llccccccc}
\toprule
\textbf{Pair} & \textbf{Mod.} & \shortstack{$s^\ast$\\ orig./rev.} & \shortstack{Mirage \%\\ orig./rev.} & \shortstack{Blackout abst.\\ orig./rev.} & \shortstack{Same\\ class} & \shortstack{Same\\ letter} & \shortstack{Conf.-correct\\ at $s=0$, rev.} \\
\midrule
APTOS & Noise & 4 / 6 & 64.8 / 100.0 & 100.0 / 0.0 & 0.0\% & 99.1\% & 85.5 \\
APTOS & Blur & 7 / 7 & 16.0 / 83.2 & 100.0 / 0.0 & 13.9\% & 85.1\% & 85.5 \\
HAM10000 & Noise & 3 / 3 & 100.0 / 93.1 & 0.0 / 0.0 & 0.0\% & 6.3\% & 39.6 \\
HAM10000 & Blur & 5 / 6 & 100.0 / 85.5 & 0.0 / 0.0 & 0.0\% & 9.9\% & 39.6 \\
VQA-RAD & Noise & 8 / 8 & 84.4 / 83.6 & 11.7 / 4.7 & 5.1\% & 5.1\% & 67.2 \\
VQA-RAD & Blur & 7 / 8 & 86.7 / 78.9 & 11.7 / 4.7 & 5.3\% & 5.3\% & 67.2 \\
\bottomrule
\end{tabular}
\end{minipage}

\noindent
\begin{minipage}{\linewidth}
\centering
\tiny
\setlength{\tabcolsep}{2pt}
\renewcommand{\arraystretch}{0.75}
\captionof{table}{MedGemma: noise draws and the rotated option order. Each noise entry is the range over four draws with different seeds, so a single value means all four draws agree, and blur is a single deterministic sweep. Values are given for the original (orig.), reversed (rev.), and rotated (rot.) option order. Mirage rates are at the original $s^\ast$, the blackout column gives the percent abstaining at $s=14$ under the reversed and rotated orders, and same class and same letter refer to the rotated run as defined for Table~\ref{tab:order}. Conf.-correct at $s=0$ gives the percent of the sampled examples that are confident-correct at $s=0$ under the reversed and rotated orders.}
\label{tab:draws}
\vspace{-2pt}
\begin{tabular}{llccccccc}
\toprule
\textbf{Pair} & \textbf{Mod.} & \shortstack{$s^\ast$\\ orig./rev./rot.} & \shortstack{Mirage \%\\ orig./rev./rot.} & \shortstack{Blackout abst.\\ rev./rot.} & \shortstack{Same\\ class} & \shortstack{Same\\ letter} & \shortstack{Conf.-correct\\ at $s=0$, rev./rot.} \\
\midrule
APTOS & Noise & 4 / 5 to 6 / 5 & 62.9 to 68.0 / 100.0 / 85.5 to 87.1 & 0.0 / 0.0 & 34.4\% & 0.4\% & 85.5 / 86 \\
APTOS & Blur & 7 / 7 / 6 & 16.0 / 83.2 / 8.6 & 0.0 / 0.0 & 81.3\% & 0.0\% & 85.5 / 86 \\
HAM10000 & Noise & 3 / 3 / 3 & 100.0 / 90.5 to 93.1 / 69.4 to 72.3 & 0.0 / 0.0 & 0.0\% & 0.1\% & 39.6 / 36 \\
HAM10000 & Blur & 5 / 6 / 6 & 100.0 / 85.5 / 94.8 & 0.0 / 0.0 & 0.0\% & 2.7\% & 39.6 / 36 \\
VQA-RAD & Noise & 8 / 8 / 12 & 82.0 to 85.2 / 82.8 to 83.6 / 96.1 to 96.9 & 4.7 / 1.6 & 35.8\% & 6.7\% & 67.2 / 75 \\
VQA-RAD & Blur & 7 / 8 / 9 & 86.7 / 78.9 / 96.1 & 4.7 / 1.6 & 43.1\% & 11.1\% & 67.2 / 75 \\
\bottomrule
\end{tabular}
\end{minipage}

\noindent
\begin{minipage}{\linewidth}
\centering
\tiny
\setlength{\tabcolsep}{2pt}
\renewcommand{\arraystretch}{0.75}
\captionof{table}{Readout validation on 119 to 120 items per pair, original (orig.), reversed (rev.), and, for MedGemma, rotated (rot.) option order (n/a: not run). Letter mass is the mean probability that the full-vocabulary softmax puts on the option letters. Agreement is the share of items for which the readout letter equals the parsed free-generation letter (n/a: no comparison available). Displayed letters give the counts of the readout letter. In the reversed order the displayed letter A is the abstention option, and in the rotated order it is C (APTOS), D (HAM10000), and B (VQA-RAD).}
\label{tab:readout-val}
\vspace{-2pt}
\begin{tabular}{llcccl}
\toprule
\textbf{Model} & \textbf{Dataset} & \shortstack{Letter mass\\ orig./rev./rot.} & \shortstack{Agreement\\ orig./rev./rot.} & \multicolumn{2}{l}{Displayed letters orig. / rev. / rot.} \\
\midrule
MedGemma & APTOS & .936 / .942 / .936 & .975 / 1.000 / 1.000 & \multicolumn{2}{l}{A23 B1 C36 D35 E25 / F25 E7 C41 D24 B23 / A37 B29 D27 F26 C1} \\
MedGemma & HAM10000 & .880 / .932 / .908 & .966 / 1.000 / .966 & \multicolumn{2}{l}{A42 E31 F20 B11 G8 C7 / C55 B32 D24 G5 F3 / B79 C16 A15 E6 G2 F1} \\
MedGemma & VQA-RAD & .995 / .992 / .994 & .983 / 1.000 / .992 & \multicolumn{2}{l}{B79 A41 / C93 B27 / C85 A35} \\
LLaVA-Med & APTOS & .410 / .510 / n/a & n/a / 1.000 / n/a & \multicolumn{2}{l}{A120 / B120 / n/a} \\
LLaVA-Med & HAM10000 & .537 / .498 / n/a & n/a / n/a / n/a & \multicolumn{2}{l}{A119 / B119 / n/a} \\
LLaVA-Med & VQA-RAD & .629 / .485 / n/a & 1.000 / .970 / n/a & \multicolumn{2}{l}{A120 / B75 C26 A19 / n/a} \\
Ovis & APTOS & .996 / .991 / n/a & 1.000 / 1.000 / n/a & \multicolumn{2}{l}{F120 / A120 / n/a} \\
Ovis & HAM10000 & .978 / .965 / n/a & n/a / 1.000 / n/a & \multicolumn{2}{l}{A119 / A119 / n/a} \\
Ovis & VQA-RAD & .986 / .986 / n/a & 1.000 / 1.000 / n/a & \multicolumn{2}{l}{C119 A1 / A118 C2 / n/a} \\
\bottomrule
\end{tabular}
\end{minipage}
\vspace{4pt}

\noindent
\begin{minipage}{\linewidth}
\centering
\tiny
\setlength{\tabcolsep}{3pt}
\renewcommand{\arraystretch}{0.75}
\captionof{table}{MedGemma: image sweeps degraded at native resolution and at the model input resolution (896 pixels, resized before degradation). Mirage rates are at each setting's own $s^\ast$. Noise rows use the original noise draw, which is separate from seeds 0 to 3. Across seeds 0 to 3 the change from native to input resolution is $-15.3$, $+5.9$, $-8.2$, and $-16.0$ points for APTOS, $-3.2$ to $-1.7$ for HAM10000, and $+4.6$ to $+7.8$ for VQA-RAD.}
\label{tab:resolution}
\vspace{-2pt}
\begin{tabular}{llcccc}
\toprule
\textbf{Pair} & \textbf{Modality} & \shortstack{$s^\ast$\\ native / input} & \shortstack{Mirage \%\\ native / input} & \shortstack{Conf.-corr. at $s{=}0$\\ input} \\
\midrule
APTOS & Noise & 4 / 3 & 64.8 / 52.7 & 99.6\% \\
APTOS & Blur & 7 / 5 & 16.0 / 24.6 & 99.6\% \\
HAM10000 & Noise & 3 / 4 & 100.0 / 97.7 & 97.7\% \\
HAM10000 & Blur & 5 / 6 & 100.0 / 100.0 & 97.7\% \\
VQA-RAD & Noise & 8 / 6 & 84.4 / 89.8 & 100.0\% \\
VQA-RAD & Blur & 7 / 8 & 86.7 / 78.1 & 100.0\% \\
\bottomrule
\end{tabular}
\end{minipage}
\vspace{4pt}

\noindent
\begin{minipage}{\linewidth}
\centering
\tiny
\setlength{\tabcolsep}{2pt}
\renewcommand{\arraystretch}{0.75}
\captionof{table}{MedGemma: commit-vs-abstain probe at $s^\ast$, nested 5-fold AUROC (layer and regularization chosen inside the training folds). Rows give the token position and whether the activations come from the image at $s^\ast$ or from the clean input. Noise conditions report the range over four noise draws (seeds 0 to 3), and blur conditions a single value, both computed on all examples. Labels come from the activation-extraction run and can differ by one example from Table~\ref{tab:mirage-rate} (VQA-RAD blur: 110 commit, 18 abstain). Class is the class-only baseline.}
\label{tab:probe-nested}
\vspace{-2pt}
\begin{tabular}{llccccccccc}
\toprule
& & \multicolumn{2}{c}{Labels} & \multicolumn{4}{c}{Activations at $s^\ast$} & \multicolumn{2}{c}{Clean input} & \\
\textbf{Pair} & \textbf{Mod.} & Commit & Abstain & \shortstack{Last\\ image} & \shortstack{Last\\ question} & \shortstack{Last\\ prompt} & \shortstack{Answer\\ prefix} & \shortstack{Last\\ prompt} & \shortstack{Best of\\ four pos.} & Class \\
\midrule
APTOS & Noise & 161 to 174 & 82 to 95 & .787 to .839 & .875 to .921 & .895 to .953 & .997 to .999 & .736 to .801 & .746 to .801 & .550 to .630 \\
APTOS & Blur & 41 & 215 & .905 & .983 & .994 & .999 & .799 & .803 & .768 \\
VQA-RAD & Noise & 105 to 109 & 19 to 23 & .396 to .593 & .851 to .921 & .966 to .970 & .991 to 1.000 & .713 to .825 & .752 to .825 & .581 to .626 \\
VQA-RAD & Blur & 110 & 18 & .818 & .747 & .896 & .991 & .810 & .810 & .576 \\
\bottomrule
\end{tabular}
\end{minipage}
\vspace{4pt}

\noindent
\begin{minipage}{\linewidth}
\centering
\tiny
\setlength{\tabcolsep}{2pt}
\renewcommand{\arraystretch}{0.75}
\captionof{table}{MedGemma: accuracy (\%) under language degradation with a separate prompt that offers no abstention option, pooled over the three severities of each attack. The first value uses one mutation per condition shared by all examples, the second is the first of four runs with an independent mutation per example (the four runs differ by at most 3.9 points for any attack and protection setting). Blackout is a single severity.}
\label{tab:lang-v2}
\vspace{-2pt}
\begin{tabular}{llcccc}
\toprule
\textbf{Dataset} & \textbf{Attack} & \texttt{protect\_all} & \texttt{protect\_desc} & \texttt{protect\_idk} & \texttt{protect\_none} \\
\midrule
APTOS & WordNet & 92.7 / 94.9 & 89.8 / 95.2 & 93.2 / 82.6 & 65.4 / 82.6 \\
APTOS & Character & 77.0 / 76.3 & 79.4 / 77.1 & 40.6 / 44.3 & 38.4 / 41.0 \\
APTOS & Blackout & 91.8 / 91.8 & 91.8 / 91.8 & 25.0 / 25.4 & 25.0 / 25.4 \\
HAM10000 & WordNet & 94.5 / 86.3 & 91.0 / 85.5 & 45.8 / 67.1 & 71.5 / 70.6 \\
HAM10000 & Character & 95.9 / 88.2 & 94.2 / 90.0 & 30.8 / 26.0 & 26.9 / 26.8 \\
HAM10000 & Blackout & 87.3 / 87.3 & 87.3 / 87.3 & 18.5 / 18.8 & 18.5 / 18.8 \\
\bottomrule
\end{tabular}
\end{minipage}
\vspace{2pt}

\textbf{Appendix K: Language Degradation, Class-Level Trajectories, and Abstention Propensity}

\textbf{Language mutation definitions.} Prompt-language information is degraded with the image held fixed at its clean form throughout. Each instruction template is decomposed into four functional components: an opening task description, an abstention instruction, answer-choice descriptions, and a closing response format. Answer-choice letters remain unmutated to preserve output-format compatibility. We then apply three attack families to eligible text: (1) \textbf{WordNet substitution:} eligible words are independently replaced by random WordNet synonyms at rate $r \in \{0.25, 0.50, 0.75\}$. (2) \textbf{Character corruption:} eligible characters are independently replaced by random lowercase letters at rate $r \in \{0.25, 0.50, 0.75\}$. (3) \textbf{Blackout masking:} every eligible word is replaced by an equal-length \texttt{X} string at rate $1.0$, forming the linguistic analogue of image blackout. We use four \emph{protection configurations} (\texttt{protect\_all}, \texttt{protect\_idk}, \texttt{protect\_desc}, and \texttt{protect\_none}) to specify whether the abstention instruction, answer-choice descriptions, both, or neither are shielded from attack. In all configurations, the opening and closing lines are mutated. This yields 28 language-degradation conditions per example, covering overall corruption severity and targeted degradation of specific prompt components. For VQA-RAD only the question text is mutated, the options and closing line are always protected, and there is a single configuration with 7 conditions.

\vspace{-4pt}
\noindent
\begin{minipage}{\linewidth}
\centering
\tiny
\setlength{\tabcolsep}{2pt}
\renewcommand{\arraystretch}{0.75}
\captionof{table}{MedGemma: VQA-RAD text degradation. Only the question text is mutated, and there is a single protection configuration.}
\label{tab:vqarad-lang-full}
\vspace{-2pt}
\begin{tabular}{lccc}
\toprule
\textbf{Attack (severity)} & \textbf{IDK\%} & \textbf{Natural acc.} & \textbf{Forced acc.} \\
\midrule
wordnet (0.25) & 0.0 & 92.2\% & 92.2\% \\
wordnet (0.50) & 0.8 & 89.1\% & 89.8\% \\
wordnet (0.75) & 1.6 & 87.5\% & 89.1\% \\
char (0.25) & 4.7 & 69.5\% & 72.7\% \\
char (0.50) & 13.3 & 43.8\% & 50.0\% \\
char (0.75) & 9.4 & 53.1\% & 55.5\% \\
blackout (1.00) & 3.1 & 50.8\% & 53.1\% \\
\bottomrule
\end{tabular}
\end{minipage}
\vspace{-4pt}

On the 64-per-class VQA-RAD population (Table~\ref{tab:vqarad-lang-full}), synonym substitution is relatively mild even at $r=0.75$ (87.5\% natural accuracy, 1.6\% IDK). Character-level corruption is substantially more damaging and non-monotonic: natural accuracy falls to 43.8\% at $r=0.50$ before recovering to 53.1\% at $r=0.75$, while IDK peaks at $r=0.50$ (13.3\%) and falls to 9.4\% at $r=0.75$. Greater linguistic corruption therefore does not monotonically increase abstention. Full blackout of the question text produces only 3.1\% IDK. Forced-commit accuracy generally exceeds natural accuracy by a few points, indicating that some IDK responses occur on examples where forced commitment recovers the correct answer.

APTOS (full $n=256$, aggregated across four protection configurations, Appendix Table~\ref{tab:aptos-lang-full-appx}) shows a similar pattern: WordNet substitution degrades accuracy gradually (94.7\% to 80.5\%) with negligible IDK, while character corruption produces a sharper drop, and character corruption and blackout masking are the only conditions eliciting meaningful abstention (10.5\% IDK at $r=0.75$, 11.2\% at blackout). On HAM10000 (full $n=346$, Appendix Table~\ref{tab:ham-lang-full-appx}) no IDK is recorded under any attack. The aggregate hides a dependence on the answer-choice descriptions: at blackout, HAM10000 accuracy is 18.5\% under \texttt{protect\_none}, 18.2\% under \texttt{protect\_idk}, 73.4\% under \texttt{protect\_desc}, and 98.8\% under \texttt{protect\_all} (Appendix Table~\ref{tab:idk-vs-noidk-appx}). With the abstention option removed, blackout accuracy is 87.3\% under both \texttt{protect\_all} and \texttt{protect\_desc} (Appendix Table~\ref{tab:lang-v2}). The non-monotonic HAM10000 WordNet accuracy (80.7, 91.3, 70.8\%) comes from one random mutation per condition pooled over four configurations, and the per-example run in Appendix J shows that shared and per-example mutation differ by up to 21 points for a given attack and protection setting, against at most 3.9 points between four per-example draws.

\textbf{Class-level trajectories.} Ground-truth answer class is associated with distinct trajectory patterns within datasets. On VQA-RAD under noise, No-answer (class B) examples are more robust than Yes-answer (class A) examples, which more often transition toward No: 34/64 No-answer examples stay correct through blackout against 3/64 Yes-answer examples, and 25/64 Yes-answer examples follow direct mirage against 2/64 No-answer examples. This pattern is consistent with the concentration of post-boundary errors on class B (Appendix Table~\ref{tab:gt-error-sstar}). On APTOS under noise, the no-DR class (A) differs markedly from the pathological grades: corrupted normal examples enter consistent abstention in 53/64 cases (83\%), whereas pathological grades (C-E) more often follow mirage-dissolve trajectories (67 to 95\%, depending on grade). HAM10000 shows similarly heterogeneous class-level trajectories: actinic keratoses (A) undergo correct recovery in 100\% of cases, while basal cell carcinoma (B), melanocytic nevi (F), and vascular lesions (G) are dominated by direct mirage (89 to 100\% under noise). Benign keratosis (C) is instead predominantly oscillatory under noise (82\%).

\vspace{-4pt}
\noindent
\begin{minipage}{\linewidth}
\centering
\tiny
\setlength{\tabcolsep}{3pt}
\renewcommand{\arraystretch}{0.75}
\captionof{table}{MedGemma: abstention behavior per pair. Mean and maximum abstention rate over $s=1,\dots,13$, abstention at blackout, share of trajectories that end in abstention, and the mirage rate at $s^\ast$.}
\label{tab:propensity}
\vspace{-2pt}
\begin{tabular}{llccccc}
\toprule
\textbf{Pair} & \textbf{Modality} & \textbf{Mean abst.} & \textbf{Max abst.} & \textbf{Blackout} & \textbf{Abst.-ending traj.} & \textbf{Mirage rate} \\
\midrule
APTOS & Noise & 62.3\% & 100.0\% & 100.0\% & 95.3\% & 64.8\% \\
APTOS & Blur & 56.6\% & 100.0\% & 100.0\% & 80.9\% & 16.0\% \\
HAM10000 & Noise & 0.1\% & 0.6\% & 0.0\% & 0.0\% & 100.0\% \\
HAM10000 & Blur & 0.6\% & 2.0\% & 0.0\% & 0.0\% & 100.0\% \\
VQA-RAD & Noise & 12.4\% & 32.8\% & 11.7\% & 12.5\% & 84.4\% \\
VQA-RAD & Blur & 17.7\% & 43.0\% & 11.7\% & 13.3\% & 86.7\% \\
\bottomrule
\end{tabular}
\end{minipage}

\textbf{Appendix L: Reproducibility and Degradation Details}\label{app:repro}

\textbf{Additive noise.} Given an image $I \in [0,1]^{H\times W\times C}$ rescaled from 8-bit integers, at severity level $s$ with coefficient $c_s$ we construct {\small\[ \textstyle \tilde{I} = \mathrm{clip}\big(I + \epsilon,\ 0,\ 1\big), \qquad \epsilon \sim \mathcal{N}\big(0,\ c_s^2\big) \mathrm{\ i.i.d.\ per\ pixel\ and\ channel}, \]} 

\noindent
before rescaling $\tilde{I}$ back to 8-bit integers for model input. Thus, $\sigma_s = c_s$ on the unit intensity scale, and the corruption is not scaled by the per-image dynamic range.

\textbf{Blur.} At severity level $s$, we convolve each channel with an isotropic Gaussian kernel of standard deviation $\rho_s$ pixels: $\tilde{I} = I * G_{\rho_s}$.

\vspace{4pt}
\noindent
\begin{minipage}{\linewidth}
\centering
\scriptsize
\setlength{\tabcolsep}{2.5pt}
\renewcommand{\arraystretch}{0.8}
\captionof{table}{Shared severity schedule for the two image degradation families. Severity 14 is a black image (BO, blackout).}
\label{tab:severity-schedule}
\vspace{2pt}
\begin{tabular}{l*{15}{c}}
\toprule
$s$ & 0 & 1 & 2 & 3 & 4 & 5 & 6 & 7 & 8 & 9 & 10 & 11 & 12 & 13 & 14 \\
\midrule
$c_s$ (noise) & 0 & .05 & .10 & .20 & .30 & .40 & .50 & .60 & .75 & .90 & 1.0 & 1.5 & 2.0 & 3.0 & BO \\
$\rho_s$ (blur, px) & 0 & .5 & 1 & 2 & 4 & 7 & 12 & 20 & 32 & 50 & 75 & 110 & 160 & 240 & BO \\
\bottomrule
\end{tabular}
\end{minipage}
\vspace{4pt}

\textbf{Abstention ablation.} At each severity level, we evaluate each example under two conditions using separate forward passes: a \emph{full} prediction using the IDK-instructed prompt, and a \emph{forced} prediction omitting both the abstention option and its instruction. We report this comparison at blackout (Appendix~C) and the entropy of committed answers in Appendix~H. Because removing the abstention option also changes the available answer space and prompt context, the comparison is interpreted as abstention-affordance sensitivity and not as causal isolation of a specific mechanism.

To characterize post-boundary commitments, we additionally perform a ground-truth and error-choice analysis. At each severity, we quantify the proportions of responses that are \emph{ground-truth}, \emph{non-ground-truth}, or \emph{abstention}. Among non-ground-truth commitments, we additionally report the distribution of predicted classes to determine whether errors concentrate on particular answers.

\raggedbottom
\subsection*{Appendix M: Natural-Correct Rate Curves, Per-Example Trajectories, Class-Level Abstention, and Trajectory Summary}

\subsubsection*{M.1 Natural-Correct Rate by Severity}
\vspace{-2pt}
\noindent
\begin{minipage}{\linewidth}
\centering
\includegraphics[width=\linewidth]{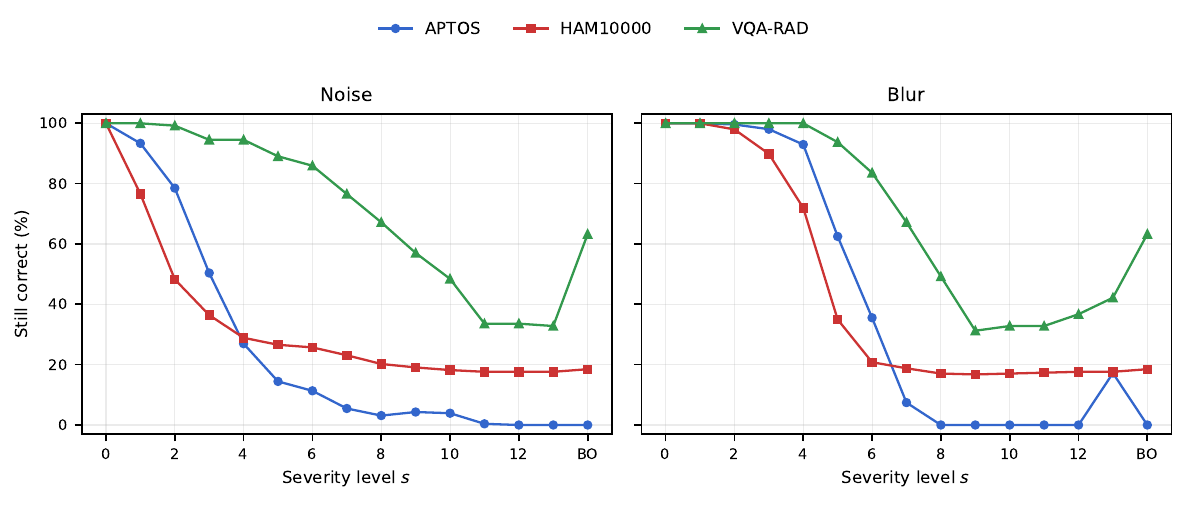}
\vspace{-2pt}
\captionof{figure}{MedGemma: natural-correct rate (\% of baseline-correct examples still correct) by severity across pairs and modalities.}
\label{fig:survival-curves}
\end{minipage}
\vspace{2pt}

\subsubsection*{M.2 Per-Example Trajectory Rasters}

\begin{figure}[H]
\centering
\begin{subfigure}[b]{0.48\textwidth}
  \includegraphics[width=\linewidth]{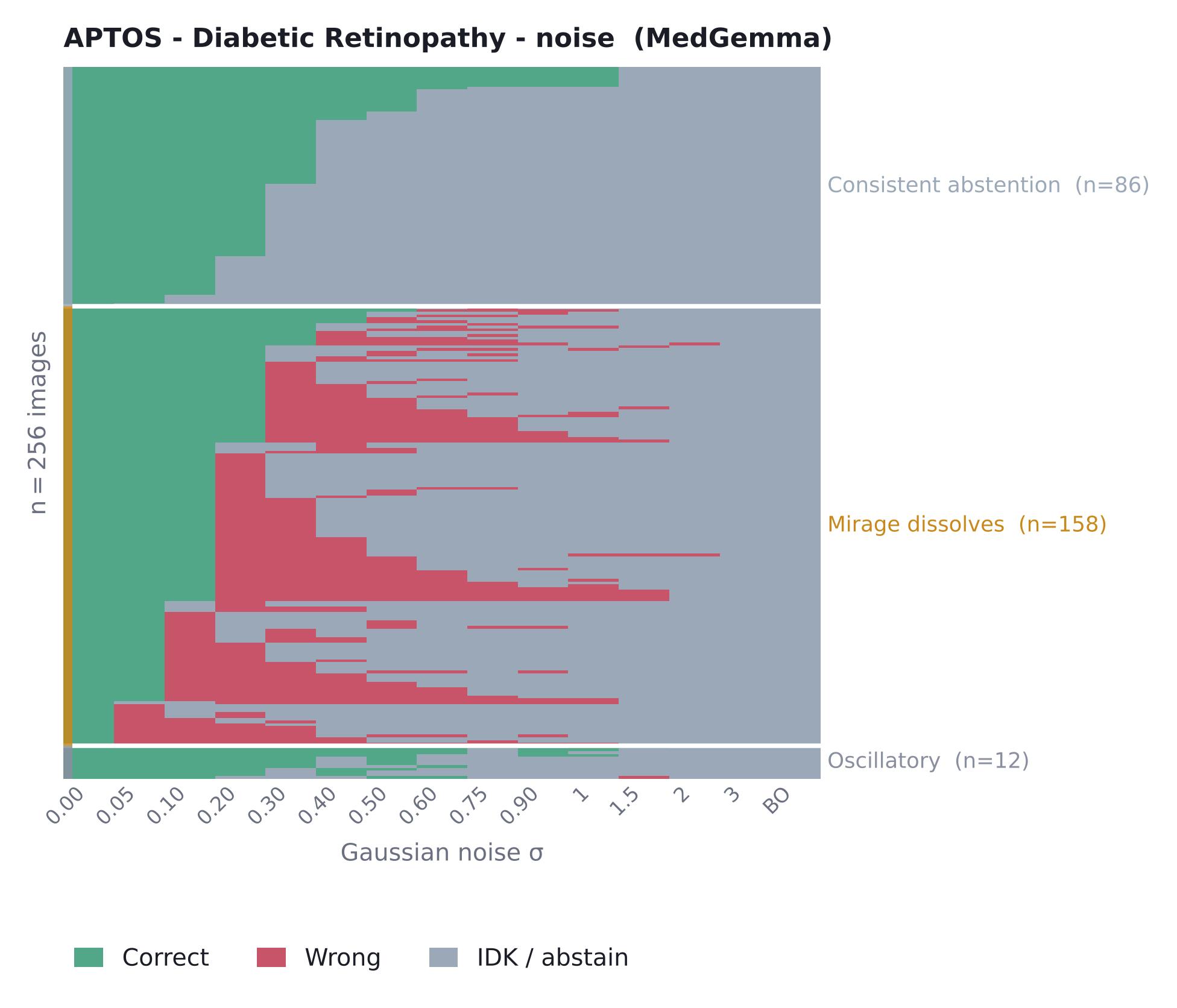}
  \caption{APTOS, noise}
\end{subfigure}
\hfill
\begin{subfigure}[b]{0.48\textwidth}
  \includegraphics[width=\linewidth]{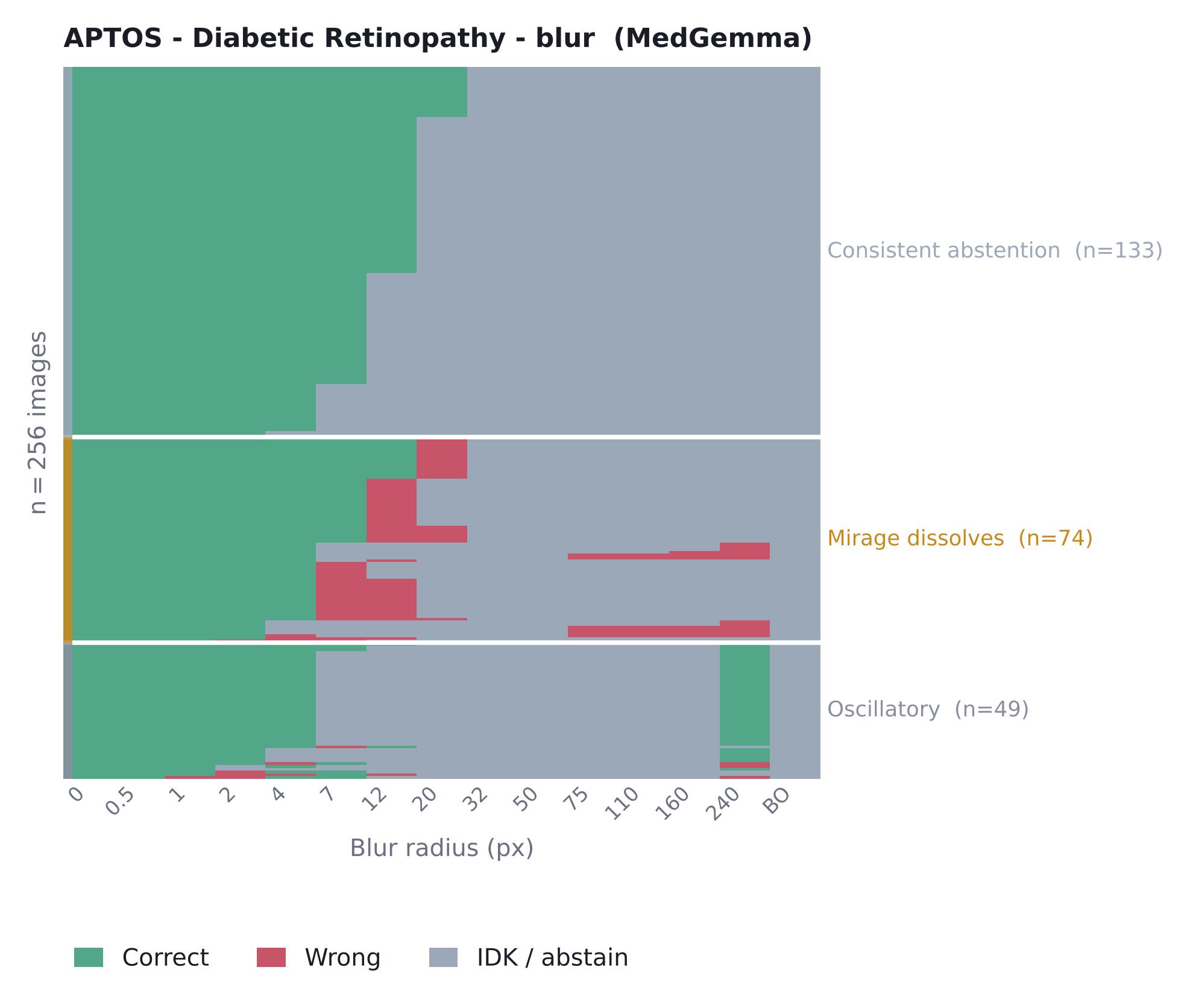}
  \caption{APTOS, blur}
\end{subfigure}

\vspace{0.5em}
\begin{subfigure}[b]{0.48\textwidth}
  \includegraphics[width=\linewidth]{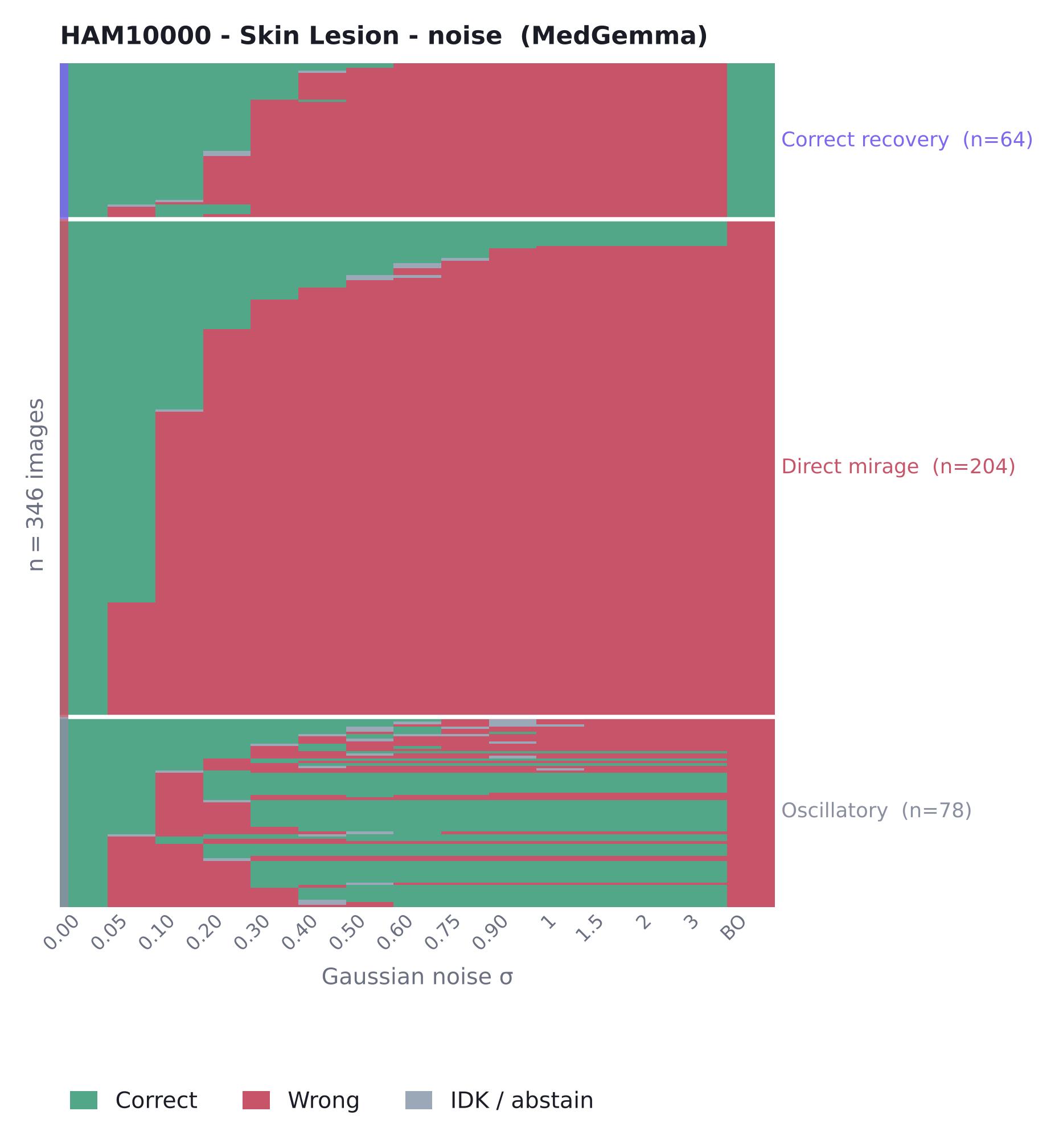}
  \caption{HAM10000, noise}
\end{subfigure}
\hfill
\begin{subfigure}[b]{0.48\textwidth}
  \includegraphics[width=\linewidth]{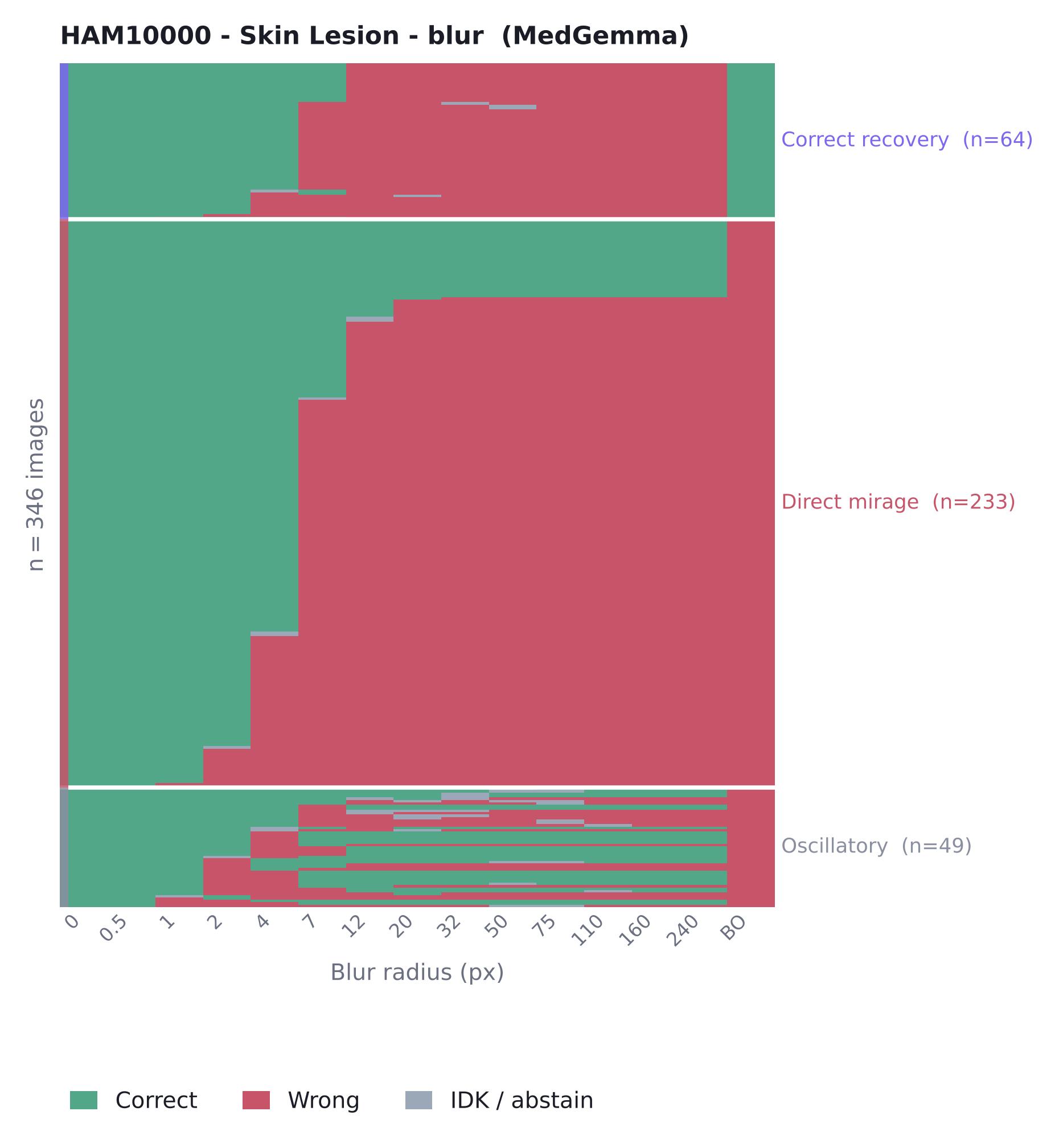}
  \caption{HAM10000, blur}
\end{subfigure}

\vspace{0.5em}
\begin{subfigure}[b]{0.48\textwidth}
  \includegraphics[width=\linewidth]{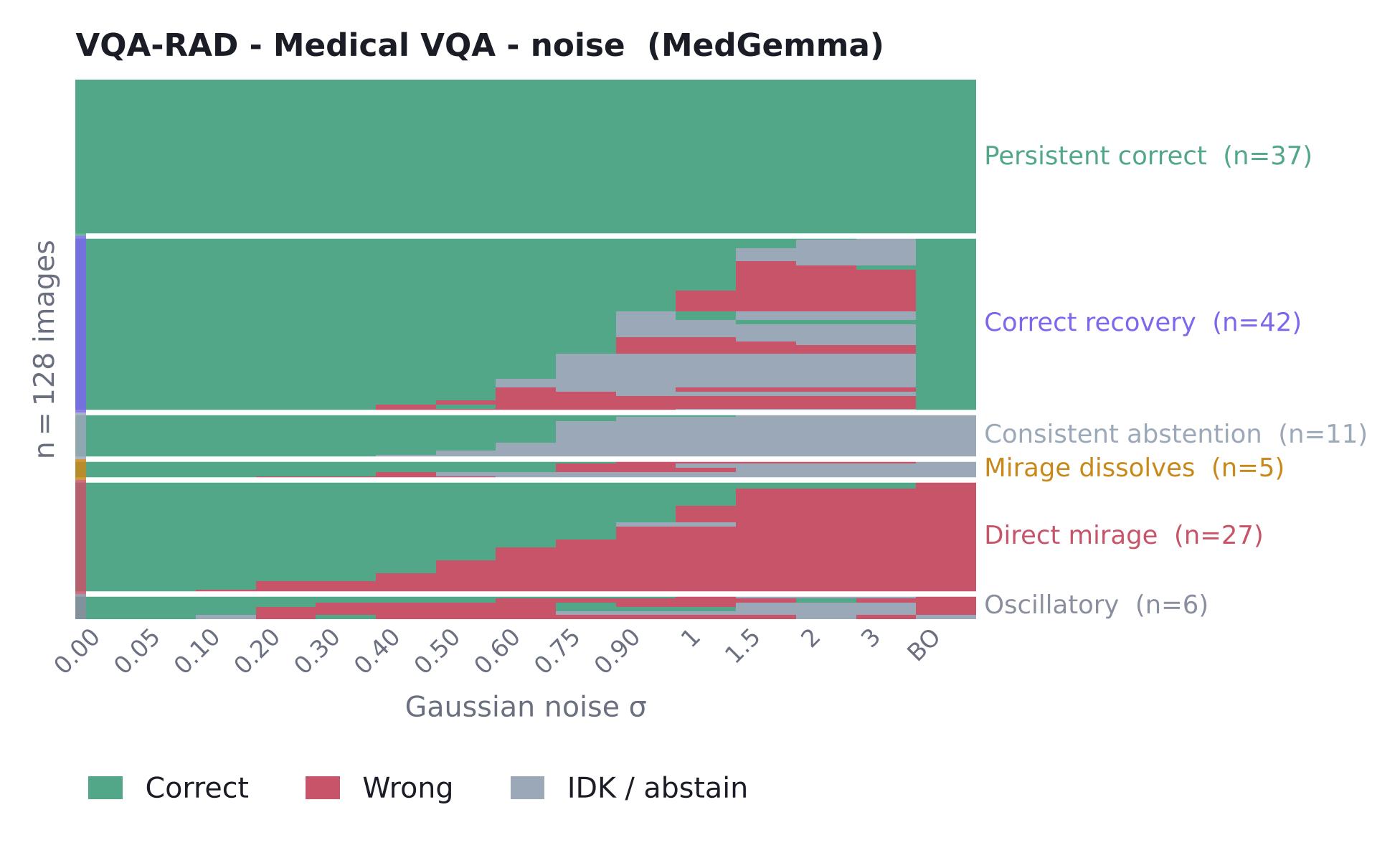}
  \caption{VQA-RAD, noise}
\end{subfigure}
\hfill
\begin{subfigure}[b]{0.48\textwidth}
  \includegraphics[width=\linewidth]{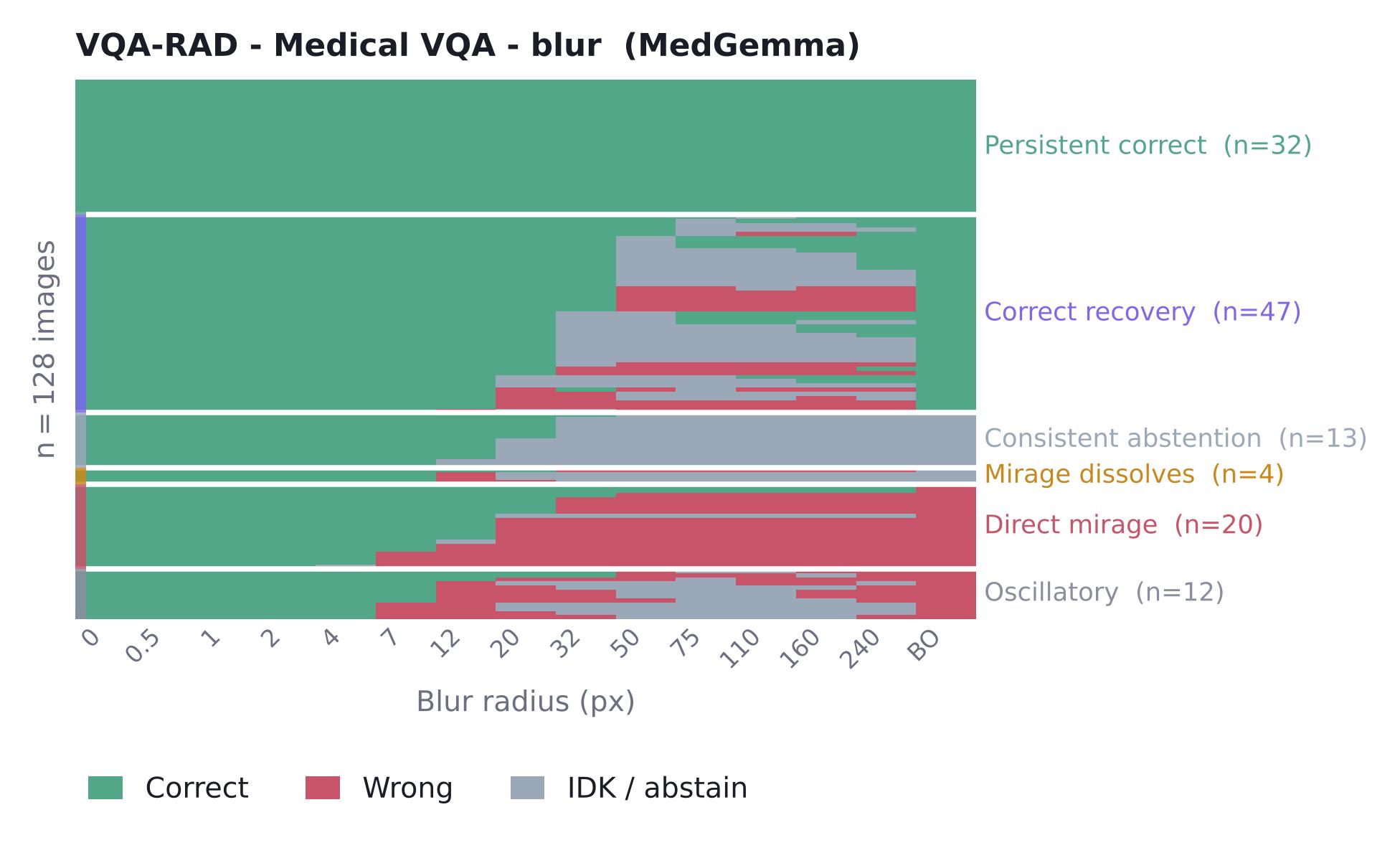}
  \caption{VQA-RAD, blur}
\end{subfigure}

\caption{Full per-example trajectory rasters across the severity schedule for every dataset-modality pair, MedGemma. Each row is one baseline-correct example, grouped by trajectory type (Persistent correct, Direct mirage, Correct recovery, Mirage dissolves, Consistent abstention, Oscillatory, as applicable to each pair). Color denotes state (correct / wrong / IDK-abstain) at each severity level. Aggregate trajectory counts underlying these panels are reported in Appendix Table~\ref{tab:trajectory-breakdown}. See Appendix~A for the trajectory taxonomy.}
\label{fig:full_rasters}
\end{figure}

\clearpage

\begin{figure}[H]
\centering

\subsubsection*{M.4 Trajectory-Type Summary}

\includegraphics[width=0.85\linewidth]{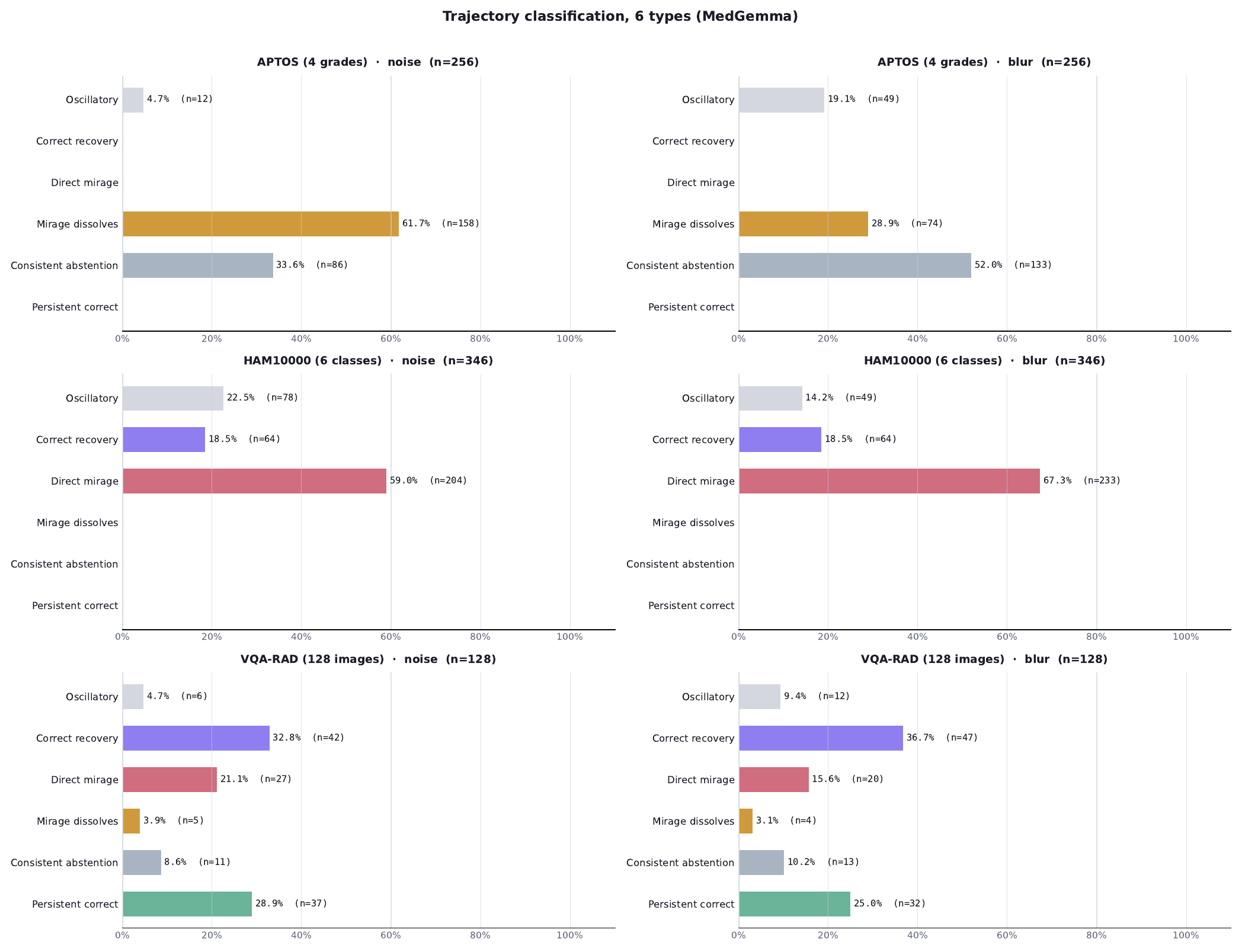}

\caption{Summary of trajectory-type proportions across all dataset-modality pairs for MedGemma. Bars aggregate the per-example rasters shown in Appendix~M.2 into trajectory-type counts, corresponding to the columns of Appendix Table~\ref{tab:trajectory-breakdown}.}
\label{fig:traj_all_bars}
\end{figure}

\end{document}